\documentclass[a4paper,11pt]{article}
\newcommand{\ReportTitle}{ZERITH~V2.5: From Predictive Power to Policy Learnability}

\newcommand{\ReportShortTitle}{ZERITH~V2.5}
\newcommand{\ReportSubtitle}{From Predictive Power\\to Policy Learnability}
\newcommand{\ReportTeam}{ZERITH Z-Lab}
\newcommand{\ReportAuthorRowBreak}{, }
\newcommand{\ReportAuthorSeparator}{, }
\newcommand{\ReportAuthorMark}[1]{\textsuperscript{\ensuremath{#1}}}
\newcommand{\ReportAuthorsPlain}{Chenxi Han, Ruby, Epoch, Orion, Luke, Bin, Dusk, Lyra, Cal, Alvin, Hubert, Alex, Jiahe Han, Tao Zhang, Hang Liu, Shoujie Li, Yizhe Li, Bokui Chen, Houde Liu, Guoping Pan, Yi Cheng}
\newcommand{\ReportAuthors}{Chenxi~Han\ReportAuthorMark{\dagger *}\ReportAuthorSeparator Ruby\ReportAuthorMark{*}\ReportAuthorSeparator Epoch\ReportAuthorSeparator Orion\ReportAuthorSeparator Luke\ReportAuthorSeparator Bin\ReportAuthorSeparator Dusk\ReportAuthorSeparator Lyra\ReportAuthorSeparator Cal\ReportAuthorSeparator Alvin\ReportAuthorSeparator Hubert\ReportAuthorSeparator Alex\ReportAuthorRowBreak Jiahe~Han\ReportAuthorSeparator Tao~Zhang\ReportAuthorSeparator Hang~Liu\ReportAuthorSeparator Shoujie~Li\ReportAuthorSeparator Yizhe~Li\ReportAuthorSeparator Bokui~Chen\ReportAuthorMark{\blacklozenge}\ReportAuthorSeparator Houde~Liu\ReportAuthorMark{\blacklozenge}\ReportAuthorSeparator Guoping~Pan\ReportAuthorMark{\dagger\blacklozenge}\ReportAuthorSeparator Yi~Cheng\ReportAuthorMark{\dagger\blacklozenge}}
\newcommand{\ReportAuthorsChinese}{Chenxi~Han\ReportAuthorMark{\dagger *}\ReportAuthorSeparator Ruby\ReportAuthorMark{*}\ReportAuthorSeparator Epoch\ReportAuthorSeparator Orion\ReportAuthorSeparator Luke\ReportAuthorSeparator Bin\ReportAuthorSeparator Dusk\ReportAuthorSeparator Lyra\ReportAuthorSeparator Cal\ReportAuthorSeparator Alvin\ReportAuthorSeparator Hubert\ReportAuthorSeparator Alex\ReportAuthorRowBreak Jiahe~Han\ReportAuthorSeparator Tao~Zhang\ReportAuthorSeparator Hang~Liu\ReportAuthorSeparator Shoujie~Li\ReportAuthorSeparator Yizhe~Li\ReportAuthorSeparator Bokui~Chen\ReportAuthorMark{\blacklozenge}\ReportAuthorSeparator Houde~Liu\ReportAuthorMark{\blacklozenge}\ReportAuthorSeparator Guoping~Pan\ReportAuthorMark{\dagger\blacklozenge}\ReportAuthorSeparator Yi~Cheng\ReportAuthorMark{\dagger\blacklozenge}}
\newcommand{\ReportAffiliation}{Zerith Robotics}
\newcommand{\ReportAffiliationChinese}{零次方}
\newif\ifReportDraft
\ReportDraftfalse

\newcommand{\ReportEmail}{business@zerith.com}
\newcommand{\ReportProjectURL}{https://www.zerith.com/blog/zerith-2-5/}

\newif\ifReportShowResources
\ReportShowResourcesfalse

\usepackage[T1]{fontenc}
\usepackage{newtxtext}
\usepackage{amsmath,graphicx,booktabs,cite,url}
\usepackage{newtxmath}
\usepackage{placeins,float,xcolor}
\newcommand{\ReportBodySize}{11.0}
\newcommand{\ReportBodyLeading}{14.6}
\newcommand{\ReportTableSize}{9.3}
\newcommand{\ReportTableLeading}{12.0}
\newcommand{\ReportTextWidth}{6.3}
\newcommand{\ReportTopMargin}{0.95}
\newcommand{\ReportBottomMargin}{0.8}
\definecolor{ReportMuted}{HTML}{5F6665}
\definecolor{ReportAccent}{HTML}{7130D1}
\definecolor{ReportInk}{HTML}{202526}
\definecolor{ReportBrandDark}{HTML}{27045C}
\newcommand{\ReportPaperSize}{a4paper}

\newcommand{\paperfigure}[2][\linewidth]{%
  \includegraphics[width=#1,keepaspectratio]{#2}}
\AddToHook{env/figure/begin}{}
\AddToHook{env/table/begin}{}
\newif\ifReportInTable
\AddToHook{env/table/begin}{\ReportInTabletrue}
\newcommand{\ReportTableTypography}{%
  \fontsize{\ReportTableSize}{\ReportTableLeading}\selectfont
  \renewcommand{\arraystretch}{1.08}}
\AddToHook{env/tabular/begin}{%
  \ifReportInTable\ReportTableTypography\else
    \fontsize{9}{11.5}\selectfont\renewcommand{\arraystretch}{1.08}\fi
  \setlength{\tabcolsep}{4pt}}
\AddToHook{env/tabular*/begin}{\ifReportInTable\ReportTableTypography\fi}
\newcommand{\method}{ZERITH~V2.5}
\newcommand{\sg}{\operatorname{sg}}
\title{\ReportTitle}
\author{\ReportAuthors}
\usepackage{xcolor,etoolbox,fancyhdr,titlesec,multicol,caption,siunitx,adjustbox,tikz}
\usetikzlibrary{calc}
\ifdefined\XeTeXversion
  \usepackage[protrusion=true,expansion=false]{microtype}
\else
  \usepackage[protrusion=true,expansion=true]{microtype}
\fi
\usepackage[type1,semibold,scaled=0.95]{inter}

\usepackage[\ReportPaperSize,textwidth=\ReportTextWidth in,top=\ReportTopMargin in,bottom=\ReportBottomMargin in,
  headheight=22pt,headsep=12pt,footskip=24pt]{geometry}
\definecolor{ReportText}{HTML}{414849}
\definecolor{ReportLavender}{HTML}{BCA4F3}
\definecolor{ReportLine}{HTML}{D9DCD7}
\definecolor{ReportHeaderLine}{HTML}{DCE3EE}
\definecolor{ReportPaper}{HTML}{F7F7F2}
\definecolor{ReportDark}{HTML}{1D2423}
\pagecolor{white}
\color{ReportText}
\newcolumntype{P}[1]{>{\raggedright\arraybackslash}p{#1}}
\usepackage[unicode,pdfpagelabels=true,plainpages=false,bookmarksnumbered=true,colorlinks=true,
  linkcolor=ReportAccent,citecolor=ReportAccent,urlcolor=ReportAccent]{hyperref}
\usepackage{bookmark}
\usepackage[nameinlink,noabbrev]{cleveref}
\crefname{figure}{Figure}{Figures}
\crefname{table}{Table}{Tables}
\crefname{equation}{Equation}{Equations}
\crefname{section}{Section}{Sections}
\crefname{subsection}{Section}{Sections}
\crefname{appendix}{Appendix}{Appendices}
\Crefname{appendix}{Appendix}{Appendices}
\crefname{subappendix}{Appendix}{Appendices}
\Crefname{subappendix}{Appendix}{Appendices}
\hypersetup{bookmarksdepth=2}
\newcommand{\ReportAppendixSection}[2]{%
  \FloatBarrier\par\Needspace{10\baselineskip}
  \refstepcounter{section}%
  \section*{Appendix \thesection\quad #1}%
  \addcontentsline{toc}{section}{Appendix \thesection\quad #1}%
  \label{#2}}
\ifReportDraft
  \usepackage{draftwatermark}
  \SetWatermarkText{DRAFT}
  \SetWatermarkScale{0.72}
  \SetWatermarkColor[gray]{0.93}
\fi
\hypersetup{pdftitle={\ReportTitle},pdfauthor={\ReportAuthorsPlain},
  pdflang=en-US,pdfcreator={LaTeX},pdfsubject={Video-Action Model (VAM) Technical Report},pdfkeywords={video-action models, VAM, video pretraining, policy learnability, robot learning}}
\newcommand{\ReportBrand}{\raisebox{-0.2\height}{\includegraphics[height=15pt,trim=67.2bp 52.8bp 67.2bp 53.28bp,clip]{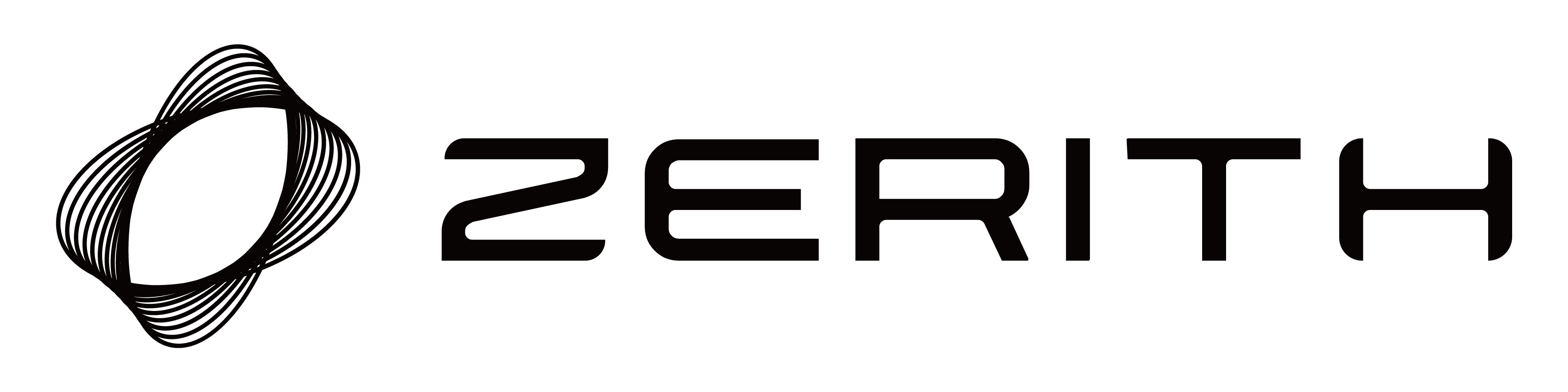}}}

\newcommand{\ReportAuthorship}{\ifdefined\ReportBilingualMode\ReportAuthorsChinese\else\ReportAuthors\fi}
\newcommand{\ReportInstitution}{\ifdefined\ReportBilingualMode\ReportAffiliationChinese\else\ReportAffiliation\fi}

\fancypagestyle{report}{%
  \fancyhf{}%
  \fancyhead[L]{\small\ReportBrand}%
  \fancyhead[R]{\fontsize{8}{11}\selectfont\sffamily\color{ReportMuted}\textcolor{ReportAccent}{\ReportShortTitle}\quad /\quad TECHNICAL REPORT}%
  \fancyfoot[C]{\fontsize{8}{10}\selectfont\sffamily\color{ReportMuted}\thepage}%
  }
\fancypagestyle{plain}{%
  \fancyhf{}%
  \fancyhead[L]{\small\ReportBrand}%
  \fancyhead[R]{\fontsize{8}{11}\selectfont\sffamily\color{ReportMuted}\textcolor{ReportAccent}{\ReportShortTitle}\quad /\quad TECHNICAL REPORT}%
  \fancyfoot[C]{\fontsize{8}{10}\selectfont\sffamily\color{ReportMuted}\thepage}%
  }
\titleformat{\section}{\normalfont\sffamily\bfseries\fontsize{14.5}{18}\selectfont\color{ReportInk}}{\textcolor{ReportAccent}{\thesection}}{0.7em}{}
\titleformat{\subsection}{\normalfont\sffamily\bfseries\fontsize{12}{15}\selectfont\color{ReportInk}}{\textcolor{ReportAccent}{\thesubsection}}{0.7em}{}
\usepackage{needspace}
\pretocmd{\section}{\Needspace{8\baselineskip}}{}{}

\titlespacing*{\section}{0pt}{18pt plus 2pt}{8pt}
\titlespacing*{\subsection}{0pt}{12pt plus 2pt}{6pt}

\fancypagestyle{reportcover}{%
  \fancyhf{}%
  \fancyfoot[C]{\sffamily\fontsize{8}{10}\selectfont\color{ReportMuted}\thepage}%
  }
\newcommand{\ReportAuthorLegend}{%
  {\fontsize{7.8}{9.8}\selectfont\color{ReportText}%
    $*$ Equal Contribution.\quad $\dagger$ Project Leader.\quad
    $\blacklozenge$ Corresponding Author.\par}}
\renewcommand{\maketitle}{%
  \thispagestyle{reportcover}%
  \pdfbookmark[0]{\ReportTitle}{report-title}%
  \par\vspace*{-35pt}
  \noindent\begin{minipage}{\linewidth}
    \raggedright
    {\sffamily\bfseries\fontsize{20}{24}\selectfont\color{ReportText}%
      \textcolor{ReportBrandDark}{\ReportShortTitle:} \ReportSubtitle\par}
    \vspace{8pt}
    {\sffamily\fontsize{9.3}{12.5}\selectfont\color{ReportText}%
      \renewcommand{\ReportAuthorRowBreak}{, }%
      \ReportAuthorship\par}
    \vspace{4pt}
    {\sffamily\fontsize{8}{10}\selectfont
      \color{ReportText}\ReportTeam\par}
    \vspace{3pt}
    \ReportAuthorLegend
    \vspace{4pt}
    {\sffamily\fontsize{8}{10}\selectfont
      Project page\quad
      \href{\ReportProjectURL}{\underline{\expandafter\nolinkurl\expandafter{\ReportProjectURL}}}\par}
  \end{minipage}\par\vspace{10pt}}

\newcommand{\ReportContentsTitle}{Contents}
\makeatletter
\patchcmd{\@dottedtocline}{\rightskip \@tocrmarg}{\rightskip \@tocrmarg plus 1fil}{}{}
\newcommand{\ReportPrintContents}{\@starttoc{toc}}
\newcount\ReportContentsSection
\newcommand{\ReportContents}{%
  \clearpage
  \pdfbookmark[0]{Contents}{report-contents}%
  \begingroup\fontsize{10.3}{14}\selectfont\hypersetup{linkcolor=ReportInk}%
  \setcounter{tocdepth}{2}\ReportContentsSection=0\relax
  \renewcommand*{\l@section}[2]{%
    \advance\ReportContentsSection by 1\relax
    \ifnum\ReportContentsSection=6\relax\columnbreak\fi
    \par\addvspace{10pt}\Needspace{4\baselineskip}%
    {\sffamily\bfseries\fontsize{10.5}{14}\selectfont\color{ReportInk}%
      \renewcommand*{\numberline}[1]{\hb@xt@\@tempdima{\textcolor{ReportAccent}{####1}\hfil}}%
      \@dottedtocline{1}{0em}{1.9em}{##1}{##2}}}
  \renewcommand*{\l@subsection}[2]{%
    \@dottedtocline{2}{0.6em}{2.7em}{##1}{##2}}
  \setlength{\columnsep}{28pt}\setlength{\multicolsep}{0pt}%
  \noindent{\sffamily\bfseries\fontsize{18}{22}\selectfont\color{ReportInk}\ReportContentsTitle\par}
  \vspace{10pt}\noindent{\textcolor{ReportAccent}{\rule{22pt}{0.45pt}}\textcolor{ReportLine}{\rule{\dimexpr\linewidth-22pt\relax}{0.35pt}}}\par\vspace{8pt}
  \begin{multicols}{2}\raggedcolumns\ReportPrintContents\end{multicols}%
  \par\endgroup\clearpage\pagenumbering{arabic}}
\makeatother
\AfterEndEnvironment{abstract}{\ifdefined\ReportBilingualMode\else\ReportContents\fi}

\makeatletter
\renewcommand\normalsize{\@setfontsize\normalsize{\ReportBodySize}{\ReportBodyLeading}
  \abovedisplayskip 10pt plus 2pt minus 4pt
  \belowdisplayskip 10pt plus 2pt minus 4pt
  \abovedisplayshortskip 0pt plus 2pt
  \belowdisplayshortskip 6pt plus 2pt minus 2pt}
\makeatother
\makeatletter
\AtBeginDocument{\immediate\write\@auxout{\string\citation{report:control}}}
\makeatother
\AtBeginDocument{\normalsize\pagenumbering{roman}}
\AtBeginEnvironment{thebibliography}{\fontsize{9.5}{11.3}\selectfont\raggedright
  \phantomsection\addcontentsline{toc}{section}{References}}
\apptocmd{\thebibliography}{\setlength{\itemsep}{0.6pt}\setlength{\parsep}{0pt}}{}{}
\ifdefined\pdfminorversion\pdfminorversion=7\pdfobjcompresslevel=2\pdfcompresslevel=9\pdftrailerid{}\fi

\DeclareCaptionFont{covercaption}{\fontsize{8.2}{10}\selectfont}

\renewenvironment{abstract}{%
  \par\begingroup\fontsize{9.5}{12}\selectfont\bfseries\vspace{3pt}\noindent{\sffamily\bfseries\fontsize{11}{14}\selectfont\color{ReportInk}Abstract}\par\vspace{4pt}\noindent\ignorespaces}
  {\par\endgroup\vspace{4pt}}

\newcommand{\ReportCreditHeading}[1]{%
  \par\Needspace{4\baselineskip}\vspace{5pt}\noindent
  {\sffamily\bfseries\fontsize{12}{15}\selectfont\color{ReportInk}#1}\par\nobreak\vspace{8pt}}
\newcommand{\ReportCreditGroup}[2]{%
  \par\Needspace{4\baselineskip}\noindent
  {\sffamily\bfseries\fontsize{10.5}{13}\selectfont\color{ReportInk}#1}\par\nobreak\vspace{3pt}%
  \noindent{\fontsize{11}{14.6}\selectfont #2}\par\vspace{8pt}}

\newcommand{\ReportLabFooter}{%
  \par\vspace{14pt}\noindent
  {\sffamily\small\bfseries\color{ReportBrandDark}\ReportTeam}%
  \quad{\sffamily\small\color{ReportMuted}\ReportInstitution}\par}

\begin{document}
\maketitle
\pagestyle{report}

\begingroup
\setlength{\intextsep}{2pt}
\captionsetup{font=covercaption,skip=3pt}
\begin{figure}[H]
\centering
\includegraphics[width=\linewidth,height=125mm,keepaspectratio]{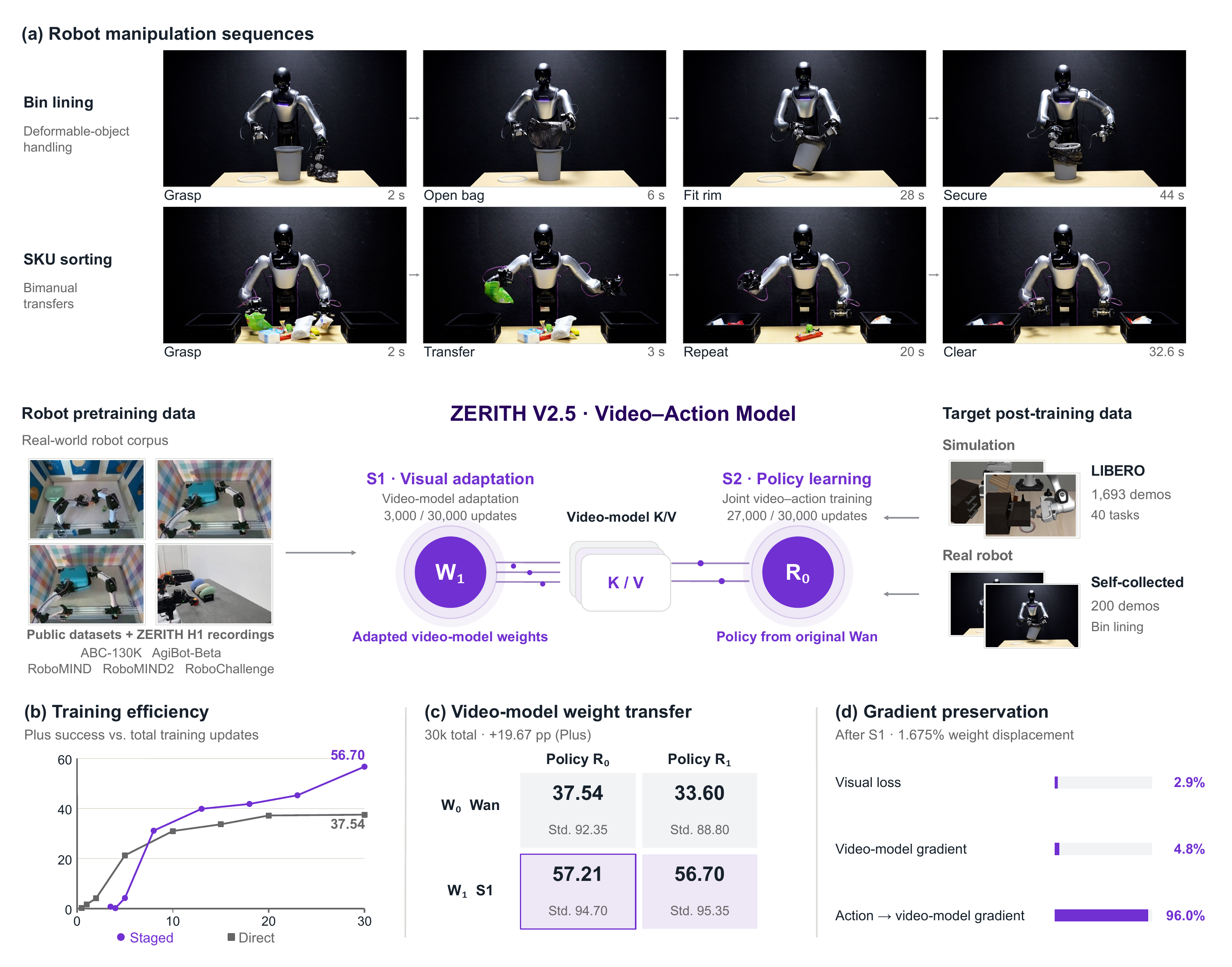}
\caption{\textbf{Adapt the visual world, then scale the action policy.} (a) Bin-lining and SKU-sorting sequences. (Center) S1 uses 3k updates for video-model adaptation, followed by 27k joint video--action updates; deployment uses $(W_1,R_0)$. Robot-data pretraining precedes this target-task schedule. (b) LIBERO-Plus learning curves: direct $(W_0,R_0)$ versus staged $(W_1,R_1)$ training. (c) Weight-transfer controls at 30k total updates; LIBERO-Plus scores appear above standard LIBERO scores. (d) Initial S2 loss and gradient measurements with $R_0$ fixed, normalized to values before S1. Subscripts 0 and 1 denote original Wan and S1-adapted weight sources, respectively.}
\label{fig:mechanism}
\end{figure}
\endgroup

\begin{abstract}
When humans learn to manipulate an unfamiliar object, they may already be able to anticipate how its state will change while still being unable to control it skillfully. This gap between prediction and control motivates us to study how the priors acquired by video generation models can be used effectively by an action policy. We present ZERITH V2.5, a video--action model (VAM) in which the action policy reads intermediate features of a video generation model, and use it to study, under fixed demonstrations and a fixed total number of updates, what kind of predictive training improves subsequent action learning.

We find that the way video predictive training is supervised affects subsequent action learning. In our experiments, adapting at many noise levels yields a large gain, while adapting only at the noise level used at deployment yields none. Weight-transfer experiments further show that the gain stays with the adapted weights when they remain in the video generation model and disappears when they are used only to initialize the action policy. Initial loss and gradient measurements at the start of joint training indicate that target-domain predictive training gives the video model a better starting state, making subsequent joint learning of prediction and action easier. Under matched data and training budgets, predictive adaptation improves control. We evaluate the complete policy on LIBERO, LIBERO-Plus, Meta-World MT50, and real-robot tasks.
\end{abstract}

\section{Introduction}

When humans learn to manipulate an unfamiliar object, they may already adjust their grip force in anticipation of upcoming load changes while their hand still deviates from the target path~\cite{flanagan2003prediction}. Prediction can develop well before skilled control. This motor-learning result suggests a question for video-driven robot learning: what training does the knowledge acquired by a video generation model need before an action policy can use it effectively?

Video generation models learn from large-scale visual data how scenes and objects change over time. Robot learning seeks to use this knowledge to help a policy generate actions from observations and task instructions. Existing methods train video prediction and action generation jointly~\cite{uva,uwm,videovla,dreamzero}, read intermediate features of the video model~\cite{vpp,fastwam,internw0}, or generate future video to support control~\cite{unipi,mimicvideo,geact2}. Predicting how images change and learning the actions needed to complete a task, however, have different training objectives. A lower video-prediction error, on its own, says little about how well an action policy can use what the model has learned. We therefore ask two questions: how should a video generation model be further trained so that it provides more useful information for action learning, and which part of the model does the resulting gain come from?

We present ZERITH V2.5 (\cref{fig:mechanism}), a video--action model (VAM) built on the Wan2.2 video generation model~\cite{wan22}, and use differences in performance across training conditions to study how video priors transfer to action ability. The video generation model computes intermediate features from the current observation; the action policy reads these features and combines them with the task instruction, the robot state, and geometric information about the current scene to generate actions. Training has two stages. Stage~1 (S1) further trains the video generation model on target-task video to predict future changes at different noise levels. Stage~2 (S2) adds action learning and optimizes the video-prediction and action-generation objectives together. Each layer of the action policy has the same structure as a layer of the video generation model, so its weights can be copied from either the original or the adapted model. This lets us vary the initialization of each and test how the gain from predictive training reaches the control task.

In controlled experiments, we fix the demonstrations and the total number of updates and compare two recipes: learning prediction and action jointly from the start (direct training), and spending the first 3,000 of 30,000 updates on video predictive training before joint learning. Across three geometric-conditioning settings with three training seeds each, the second recipe raises LIBERO-Plus success by 17.55 to 19.16 percentage points (\cref{sec:factorial}). The advantage holds at every evaluated budget from 30,000 to 50,000 updates, and direct training at 50,000 updates still scores below the staged recipe at 30,000 (\cref{sec:res-budget}).

To find where this gain comes from, we vary the initializations of the video generation model and the action policy separately. When the video generation model uses the weights after target-domain predictive training while the action policy is still initialized from the corresponding layers of the original pretrained model, LIBERO-Plus success is 19.67 points higher than with direct training. Conversely, when the video generation model starts from the original weights and only the action policy is initialized from the adapted layers, success is 3.94 points below direct training (\cref{sec:transfer}). These results suggest that the control gain comes mainly from adapting the video generation model to the target domain rather than from the action policy's initialization.

The noise setting of predictive training also matters. Training across a range of noise levels gives higher control success. Holding the future inputs at pure noise throughout matches the input seen at deployment, yet success is 20.80 points lower (\cref{sec:res-noise}). This suggests that the range of noise levels covered in training matters, and that matching the deployment condition is not enough. Two factors therefore shape the outcome: the predictive supervision the video generation model receives, and the module that receives the adapted weights. A further control shows that action supervision remains useful to the video generation model once joint learning begins: blocking action gradients into it throughout direct training lowers success below standard direct training.

To explain these differences, we analyze the predictive objective and the state of the model when action learning begins. A theoretical analysis shows that, for the predictive objective used here, different noise levels impose different predictive constraints (\cref{sec:theory-noise}). With the action-policy initialization held fixed, target-domain predictive training reduces the prediction loss of the video generation model 33.9-fold and its gradient norm by 95.2\%, while the gradient that the action loss sends into the video model keeps 96.0\% of its magnitude (\cref{sec:probes}). We hypothesize that predicting the target scene well before action learning begins gives joint training a better starting point, and eases the burden of adapting both models to the target task at once.

At deployment, the action policy reads intermediate features of the video generation model directly, without fully decoding future video frames, and the features from one forward pass are reused across all steps of the action solver. To evaluate the complete system, we pretrain on 10,000 hours of cleaned and selected real-robot data and then adapt to downstream tasks. A single checkpoint of the complete system reaches 98.95\% on LIBERO and 93.70\% on LIBERO-Plus (\cref{sec:res-benchmarks}), and the complete policy reaches 92.3\% on Meta-World MT50. On a real robot, it succeeds in 97 of 100 pick-and-place trials and 8 of 20 bin-lining trials, matching or exceeding $\pi_{0.5}$ and Fast-WAM on every task (\cref{sec:res-real}). These system evaluations test the method as a whole, while the controlled experiments with fixed data and update budgets identify the training conditions under which predictive priors transfer.

\noindent\begin{minipage}{\linewidth}
In summary, this report makes three contributions:
\begin{itemize}
\item \textbf{Training conditions for transferring video priors to control.} With fixed demonstrations and a fixed total number of updates, we compare training recipes, verify the control gain of target-domain video predictive adaptation, and examine the gain across geometric-conditioning settings and training budgets (\cref{sec:res-static,sec:factorial,sec:res-crossover,sec:res-budget}).
\item \textbf{Source of the adaptation gain and the optimization starting point.} Branch-initialization and noise-coverage controls separate the role of where the adapted weights are used from the role of the predictive supervision. Together with a theoretical analysis and diagnostics of initial losses and gradients, they offer mechanistic clues to the differences in subsequent action learning (\cref{sec:transfer,sec:res-noise,sec:probes,app:theory}).
\item \textbf{A robot control system built on reusable predictive memory.} The video predictive representation, current-scene geometry, and action readout are organized as separate modules, and one pass of the video branch supports a multi-step action solve. With pretraining on 10,000 hours of real-robot data, the complete system is evaluated on simulation benchmarks and real-robot tasks (\cref{sec:system,sec:results}).
\end{itemize}
\end{minipage}

\section{Model Architecture}
\label{sec:system}

\subsection{Overview}
\begin{figure}[!htbp]
\centering
\paperfigure[\linewidth]{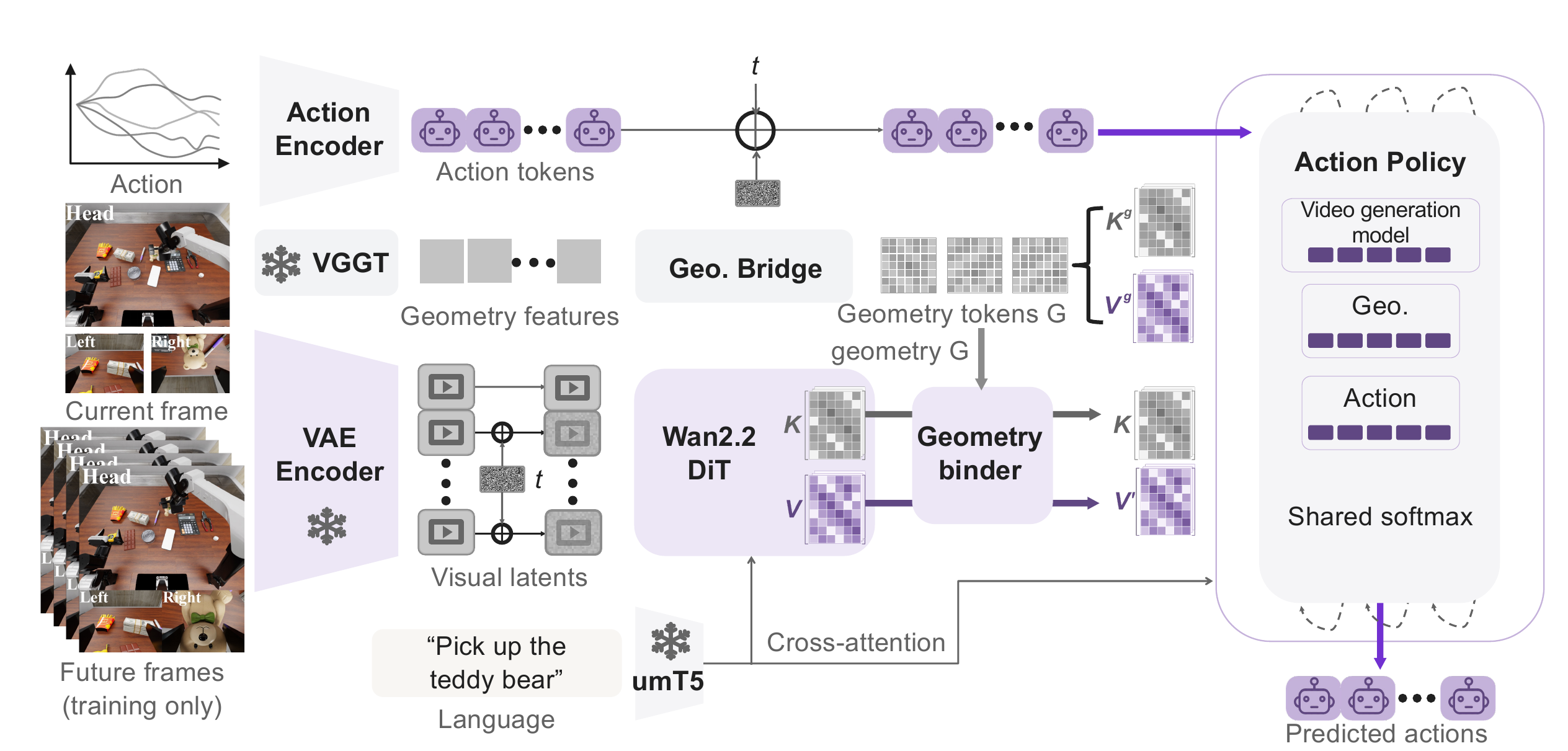}
\caption{\textbf{Video--action architecture in Stage~2 and at deployment.} Frozen encoders supply visual latents, language embeddings, and current-frame geometric features. Clean future frames (lower left) provide velocity targets during training only, whereas future latent slots contain pure noise during Stage~2 and at deployment. The action policy fuses exported video-model memory, static geometry tokens, and action queries through a single shared-softmax attention layer. Stage~1 adapts only the video generation model; Stage~2 attaches the action policy and geometry pathway. In the branch-initialization study, the video generation model and action policy are initialized independently from original or adapted weights.}
\label{fig:architecture}
\end{figure}

ZERITH V2.5 keeps the video generation model and the action policy as separate modules that communicate only through exported attention keys and values, which is what makes the staged recipe of \cref{sec:method} possible. Because this interface decouples visual dynamics modeling from action readout, the video generation model can first be adapted alone to ground its representation in the robot's visual domain before any policy parameters are introduced. At both Stage~2 and deployment, a single forward pass of the video generation model supplies the memory queried by the action policy, without fully decoding future video frames. Moreover, because each policy block mirrors the transformer architecture of its source video-model block, the two branches can be initialized independently from either original or adapted video weights (\cref{sec:transfer}). At each decision step, ZERITH V2.5 takes images from up to three camera views, a natural-language instruction, and the robot's proprioceptive state to predict a 16-step action chunk, executing a configurable number of steps before replanning; the reported deployment executes the first four (\cref{fig:architecture}).

Perceptual inputs are encoded by frozen pretrained networks: the Wan2.2 video autoencoder extracts visual latents, its umT5 encoder embeds the language instruction, and the VGGT aggregator~\cite{vggt} provides current-frame geometric features. Three trainable modules then map these inputs to control: a 30-block Wan2.2-TI2V-5B video generation model~\cite{wan22} that processes the current visual latent alongside four future latent slots, a 10-block flow-matching action policy~\cite{flowmatching} that attends to keys and values exported from ten video-model blocks, and a lightweight geometry pathway that projects current-frame VGGT features into static geometry tokens and a multiplicative residual on the exported future values. The trainable model contains approximately 7B parameters.

\subsection{Attention in the Video Generation Model}
Each camera view is encoded independently and placed into a designated slot within a single composite latent frame, distinguished by a learned camera embedding. The video generation model processes the current frame ($h=0$) together with four future latent slots under a frame-causal attention mask: tokens in the current frame attend only to the current frame, while tokens in each future slot attend to their own frame and all preceding frames. Within any given frame, tokens from different cameras attend to one another without restriction, allowing every viewpoint to be contextualized by the others (\cref{sec:camera} quantifies the contribution of this cross-camera interaction). For fixed model weights, timestep conditioning, text, and attention masks, this block-triangular mask makes the current-frame representation independent of the contents of future slots. Replacing only those contents by pure Gaussian noise therefore leaves the current-frame features unchanged. At deployment, the video generation model executes a single forward pass per decision and terminates after block 24, the deepest layer tapped by the action policy; the final five blocks and the visual-velocity output head are needed for the visual-loss computation and are omitted from the deployed forward pass.

\subsection{Action Policy}
The action policy extracts self-attention keys and values from ten fixed video-model blocks at the current frame ($h=0$) and future horizons $h\in\{1,2,4\}$. Using zero-based indices, it taps blocks 9 through 14 together with blocks 16, 18, 23, and 24, sampling densely across the middle of the backbone and more sparsely toward the output. The choice follows a design rationale: early layers primarily capture low-level appearance before mid-level scene structure emerges, whereas the final layers specialize in reconstructing the latent-space denoising target. In controlled ablations (\cref{sec:res-interface}), the point estimates for this middle-to-late profile are 0.15 to 0.20 percentage points above those for uniformly spaced layers and the final ten layers; we keep the selected indices fixed across all experiments.

Let $W_0$ denote the pretrained Wan2.2 video-model weights and $W_1$ the weights obtained after Stage~1 visual adaptation. Because each policy block is structurally identical to a video-model block, its transformer weights can be copied directly from the corresponding layers of the video generation model: $R_0$ initializes the policy from $W_0$, and $R_1$ initializes it from $W_1$. This symmetry decouples the initialization of the action policy from that of the video generation model, enabling the $2\times2$ branch-initialization study in \cref{sec:transfer} across all four $(W_i,R_j)$ pairs. The geometry pathway, proprioceptive and action embeddings, and output heads are freshly initialized in every condition. The inherited transformer weights are selected independently for the two branches; the training schedules and budgets for the four combinations are specified in \cref{sec:transfer}.

Within each policy block, action queries attend jointly to exported video-model memory, static geometry tokens, and action tokens through a single shared-softmax attention operator:
\begin{equation}
\operatorname{Attn}_a=\operatorname{softmax}\!\left(\frac{Q_a[K_w;K_g;K_a]^\top}{\sqrt{128}}+M\right)[V'_w;V_g;V_a],
\label{eq:attn}
\end{equation}
where $Q_a$ denotes the action queries, $K_w,K_g,K_a$ and $V'_w,V_g,V_a$ denote the keys and values from the video-model, geometry, and action streams, brackets indicate token-wise concatenation, 128 is the per-head dimension, $M$ masks padded or invalid tokens, and $V'_w$ denotes the geometrically modulated video-model values. Unifying all three streams under a single softmax differs from summing separate cross-attention layers: video-model, geometry, and action tokens compete directly for a shared probability mass, making the policy's allocation of attention across modalities an explicit, measurable quantity (\cref{sec:res-attention}). Following the self-attention layer, each policy block applies cross-attention to the frozen umT5 instruction embedding followed by a feed-forward network, with both self-attention and feed-forward layers modulated by the flow-matching timestep via adaptive layer normalization. The robot's proprioceptive state is projected and added to every action token, and action tokens are aligned with the temporal axis of the video latents according to physical time, placing four control steps between consecutive latent frames (\cref{sec:res-interface} compares physical-time indexing against step-index encoding). Because the exported video-model keys and values depend only on the visual inputs and instruction, a single forward pass of the video generation model supplies memory for all ten Euler integration steps of the flow-matching action solver, amortizing the cost of the 5.00B-parameter backbone across the entire action chunk.

\subsection{Geometric Conditioning}
Geometric conditioning supplements the exported video-model features with current-scene 3D features. Its interaction with staging and the choice of operator are evaluated separately in \cref{sec:factorial,sec:res-interface}. The geometry pathway comprises a 6.3M-parameter geometry bridge and a 23.6M-parameter multiplicative binding module. For each camera view, the frozen VGGT aggregator extracts current-frame features pooled to a $7\times7$ spatial grid; the geometry bridge applies layer normalization, a linear projection, a learned camera embedding, and a second layer normalization to map these features into geometry tokens $G$ aligned with the video latent patches. Each policy block projects $G$ with the key and value projections of its own self-attention layer to form static geometry memory $(K_g,V_g)$. In parallel, the binding module injects a low-rank multiplicative residual into the exported future video-model values. For the tapped horizons, define consecutive temporal differences $\Delta V_1=V_1-V_0$, $\Delta V_2=V_2-V_1$, and $\Delta V_4=V_4-V_2$, omitting the block index $\ell$ for brevity. At policy block $\ell$ and future horizon $h\in\{1,2,4\}$, the modulated video-model values are computed as
\begin{align}
g_\ell&=W^g_\ell\operatorname{LN}(G),& d_{\ell,h}&=W^d_\ell\operatorname{LN}(\Delta V_h),\nonumber\\
r_{\ell,h}&=W^o_\ell(g_\ell\odot d_{\ell,h}),& V'_{\ell,h}&=V_{\ell,h}+\tanh(\alpha_{\ell,h})r_{\ell,h},
\label{eq:binding}
\end{align}
where $\operatorname{LN}$ denotes non-parametric layer normalization and $\odot$ denotes elementwise multiplication. The bias-free matrices $W^g_\ell$ and $W^d_\ell$ project both streams into a 256-dimensional bottleneck, and $W^o_\ell$ projects the product back to the video-model value dimension. By construction, \cref{eq:binding} satisfies four algebraic properties. First, because every scalar gate $\alpha_{\ell,h}$ is initialized to zero, the residual vanishes at initialization and preserves the inherited video-model values intact. Second, the bilinear product $g_\ell\odot d_{\ell,h}$ vanishes whenever either factor is zero, ensuring that static geometry only modulates predicted visual change $\Delta V_h$ rather than injecting a static offset. Third, the update is constrained to the column space of $W^o_\ell$, a subspace of rank at most 256. Fourth, video-model keys $K_w$, current-frame values $V_0$, and the internal states of the video generation model remain untouched, leaving the backbone's own predictions and attention patterns invariant. Our conditioning ablations (\cref{sec:res-interface}) replace $g_\ell\odot d_{\ell,h}$ with alternative coupling functions $F(g,d)$; setting $F=0$ retains only the static geometry memory $(K_g,V_g)$, whereas the no-geometry baseline removes $(K_g,V_g)$ as well.

\section{Training Recipe}
\label{sec:method}

Both stages use conditional flow matching~\cite{flowmatching}. For clean data $x^\star$ and independent Gaussian noise $\epsilon$, define $x_\sigma=(1-\sigma)x^\star+\sigma\epsilon$ and target velocity $v^\star=\epsilon-x^\star$. The noise level $\sigma$ ranges from zero for clean data to one for pure noise. Losses include only valid elements, excluding padded visual patches and action dimensions.

\subsection{Stage 1: Visual Adaptation}
S1 keeps the current-frame latent clean and trains the video generation model to predict velocities for valid future visual patches. The four future slots share one noise level, drawn as $\sigma_w=0$ with probability 0.1 and $\sigma_w\sim\mathrm{Unif}(0,1)$ otherwise, while the Gaussian noise itself is sampled independently for each slot. At $\sigma_w=0$ the future inputs are clean and the target remains the velocity $\epsilon-x^\star$. S1 trains the video generation model alone: the action policy and the geometry pathways do not yet exist. We refer to this distribution as the broad-noise schedule.

The schedule is intentionally broader than the operating point at which the backbone is later read. S2 and deployment evaluate the video generation model only at $\sigma_w=1$, yet S1 covers the entire interval from clean to fully noised. The reason is that different noise levels supervise different aspects of the future: the population-optimal output at the fully noised endpoint depends on the conditional mean, whereas intermediate levels depend on posteriors across a range of signal-to-noise ratios (\cref{sec:theory-noise}). The fixed-noise control of \cref{sec:res-noise} compares this schedule directly against adaptation at the deployed endpoint.

\subsection{Stage 2: Joint Training}
S2 initializes the video generation model from $W_1$ and attaches the action policy, initialized from the selected original Wan layers ($R_0$) in the deployed configuration, or from the corresponding S1-adapted video-model layers ($R_1$) in the corresponding experimental variants. Future visual inputs are held at $\sigma_w=1$, the same condition the deployed system encounters, so a single video-model forward pass yields both the visual-velocity predictions that supervise the video generation model and the features that the policy consumes. Training and deployment thereby share one forward computation, and neither fully decodes a future frame. The action sequence receives an independently sampled noise level $\sigma_a\sim\mathrm{Unif}(0,1)$, shared across all 16 action steps. The joint objective is
\begin{equation}
\mathcal L=\mathcal L_w+\mathcal L_a,
\quad \mathcal L_a=\mathbb E\!\left[\|\hat v_a-(\epsilon_a-a^\star)\|^2_{\mathrm{valid}}\right],
\label{eq:loss}
\end{equation}
where $a^\star$ is the clean action sequence, $\epsilon_a$ is Gaussian action noise, $\hat v_a$ is the policy's velocity prediction, $\mathcal L_w$ is the analogous masked visual loss, and the valid-element norm excludes padded dimensions. The visual loss is retained throughout S2 with unit weight. The adapted video generation model continues to receive visual supervision throughout S2 alongside the action objective. Video-model parameters receive visual-loss gradients at every step and action-loss gradients according to the routing schedule below.

\subsection{Gradient Routing Between the Branches}
For a video-feature tensor $X$ supplied to the policy, we apply
\begin{equation}
\widetilde X=\sg(X)+\beta\,(X-\sg(X)),
\label{eq:grad}
\end{equation}
where $\sg$ denotes stop-gradient. \cref{eq:grad} is the identity in the forward direction for every $\beta$ and scales the returned gradient by $\beta$. As a function of normalized S2 progress $p\in[0,1]$, we set $\beta=0$ until $p=0.05$, increase it linearly to one by $p=0.10$, and hold it at one thereafter. This fixed schedule initially trains the video generation model on its visual loss alone, allowing the newly initialized policy to adapt to its inputs. Action gradients are introduced after the prescribed initial interval and ramped to full strength; the schedule does not detect when those gradients become task-relevant. The video generation model receives visual-loss gradients throughout. The full-isolation control holds $\beta=0$ for the entire run of direct training.

\subsection{Budget Allocation and Crossover Model}
\label{sec:discussion}
The standard endpoint controls compare schedules at 30,000 total downstream updates on the same demonstrations. In this comparison, staged training allocates the first 3,000 to S1 and the remaining 27,000 to S2; direct training allocates all 30,000 to joint updates. The two schedules use the same total number of updates, separating the effect of the staged recipe from that of additional training. The recipe differs from direct training in its ordering, its S1 noise distribution, and the absence of the policy during S1. Because S1 omits the action policy and the geometry pathways, it processes fewer input tokens per update than joint training, and \cref{sec:tokens} quantifies the resulting reduction in action-token processing. Learning curves also report matched S2 updates, budget sweeps vary total updates, and additional robot-data pretraining is accounted for separately.

\begin{figure}[!htbp]
\centering
\paperfigure[0.58\linewidth]{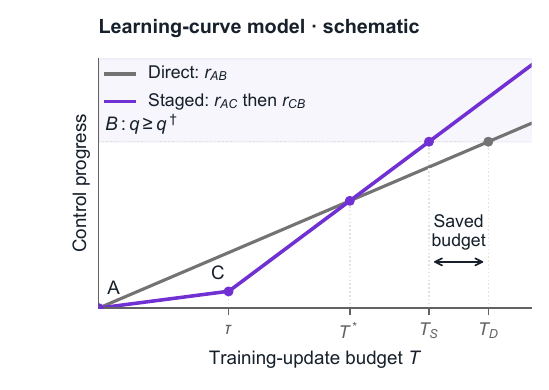}
\caption{\textbf{Illustrative training-budget crossover.} Normalized schematic of \cref{eq:curves}. $A$ is the pretrained state, $C$ the S1-adapted state, $B=\{q\ge q^\dagger\}$ the target-performance region, and $r_{AB},r_{AC},r_{CB}$ the progress rates along the direct, S1, and subsequent S2 routes. $\tau$ is the S1 budget and $T^*$ the equal-progress crossing. At the target progress $q^\dagger$, $T_S$ and $T_D$ are the staged and direct budgets; the arrow marks the saving $T_D-T_S$.}
\label{fig:crossover}
\end{figure}

To make the budget trade-off explicit, let $A=(W_0,R_0)$ denote the pretrained initialization, $C=(W_1,R_0)$ the state after visual adaptation, and $B$ a target-performance region; direct training follows $A\to B$ and staged training follows $A\to C\to B$. \cref{sec:theory-area} states the crossover condition without assuming a functional form; here we use a closed-form example in which $T$ and $\tau$ count optimizer updates and let $q(T)$ denote performance measured by a common probe that is defined throughout both schedules, including video-model-only S1. Assume constant progress rates within each stage. Direct training starts at $q_0$ with rate $r_{AB}$; staged training uses rate $r_{AC}$ for budget $\tau$, followed by rate $r_{CB}$:
\begin{align}
q_D(T)&=q_0+r_{AB}T,\nonumber\\
q_S(T)&=\begin{cases}
q_0+r_{AC}T,&T\leq\tau,\\
q_0+r_{AC}\tau+r_{CB}(T-\tau),&T>\tau.
\end{cases}
\label{eq:curves}
\end{align}
Assume $0\leq r_{AC}<r_{AB}<r_{CB}$: video-model adaptation initially advances the common control metric more slowly, then provides a higher rate during joint learning. For $\tau>0$ and rates constant over the interval, equating the two curves gives
\begin{equation}
T^*=\frac{r_{CB}-r_{AC}}{r_{CB}-r_{AB}}\,\tau>\tau.
\label{eq:crossover}
\end{equation}
Under these assumptions, direct training has higher performance for $0<T<T^*$, and staged training has higher performance for $T>T^*$ (\cref{fig:crossover}). The additional budget after S1 is
\begin{equation}
T^*-\tau=\frac{(r_{AB}-r_{AC})\tau}{r_{CB}-r_{AB}}.
\label{eq:payback}
\end{equation}
The numerator is the progress forgone during S1 and the denominator is the subsequent rate advantage; a longer adaptation stage therefore recovers its cost within the same budget only if it yields a correspondingly larger S2 advantage. For fixed rates, $T^*$ increases linearly with $\tau$ and diverges as $r_{CB}$ approaches $r_{AB}$ from above. In this model, the return on staging depends jointly on the adaptation budget and the subsequent progress rates.

\cref{tab:notation} collects the notation used in the remainder of the report.

\begin{table}[!htbp]
\centering
\caption{\textbf{Notation.}}
\label{tab:notation}
\begin{tabular}{@{}lp{0.72\linewidth}@{}}
\toprule
\textbf{Symbol} & \textbf{Meaning}\\
\midrule
S1 & Stage 1: video-model-only visual adaptation on target-domain video across the full noise range (\cref{sec:method})\\
S2 & Stage 2: joint training of the video generation model and action policy, with future slots at $\sigma_w=1$\\
Direct training & Joint training from the first update, using the full budget\\
$W_0$, $W_1$ & Original Wan video-model weights; S1-adapted video-model weights\\
$R_0$, $R_1$ & Policy blocks copied from $W_0$; copied from $W_1$\\
$\sigma_w$ & Noise level of the future visual slots, from 0 (clean) to 1 (pure noise)\\
$\beta$ & Gradient-routing coefficient on the action gradient into the video generation model (\cref{eq:grad})\\
$\tau$ & Adaptation budget; 3,000 updates in the staged schedule\\
$A$, $C$, $B$ & Pretrained state $(W_0,R_0)$, adapted state $(W_1,R_0)$, target-performance region\\
$r_{AB}$, $r_{AC}$, $r_{CB}$ & Progress rates of direct training, of S1, and of S2 after adaptation\\
$T^*$ & Crossover budget in optimizer updates at which staged and direct performance are equal (\cref{eq:crossover})\\
Points (pp) & Differences between success rates, in percentage points\\
\bottomrule
\end{tabular}
\end{table}

\section{Data and Experimental Setup}
\label{sec:data}

\subsection{Robot Pretraining Data}
The pretraining corpus contains 10,000 hours of real-robot recordings selected after cleaning: 6,000 hours of proprietary data and 4,000 hours of public data (\cref{tab:data-composition}). Proprietary data comprise our ZERITH H1 robot recordings. Public sources comprise ABC-130K~\cite{abc130k}, RoboMIND 2.0~\cite{robomind2}, AgiBot World Beta~\cite{agibot}, and RoboChallenge~\cite{robochallenge}. This corpus is used for the robot-data pretraining of the complete system in \cref{sec:results}. The controlled experiments of \crefrange{sec:staging}{sec:analysis} train on the downstream demonstrations of \cref{sec:downstream}, except for the mixed-data variants of \cref{sec:res-corpus}; their starting points are listed in \cref{sec:setup}. \cref{app:data} reports processing and quality control for each source.

\begin{table}[htb]
\centering
\caption{\textbf{Selected real-robot pretraining data after cleaning.} The selected corpus comprises 6,000 hours of ZERITH H1 data and 4,000 hours of public data, totaling 10,000 hours. Cleaning and selection are described in \cref{app:data}.}
\label{tab:data-composition}
\small\setlength{\tabcolsep}{5pt}
\begin{adjustbox}{max width=\linewidth}
\begin{tabular}{P{.18\linewidth}P{.60\linewidth}r}
\toprule
\textbf{Data category} & \textbf{Sources} & \textbf{Selected hours}\\
\midrule
Proprietary & Our ZERITH H1 robot recordings & 6,000\\
Public & ABC-130K; RoboMIND 2.0; AgiBot World Beta; RoboChallenge & 4,000\\
\midrule
Total & & 10,000\\
\bottomrule
\end{tabular}
\end{adjustbox}
\end{table}

\subsection{Downstream Data and Benchmarks}
\label{sec:downstream}
LIBERO downstream training uses 1,693 demonstrations from the OpenVLA-modified LIBERO dataset~\cite{libero,openvla}, covering 40 tasks; every schedule and initialization draws from this same set. The smaller dataset used in the adaptation ablation contains 13,259 ABC-130K episodes and 1,693 LIBERO demonstrations; it is identified as the mixed dataset in \cref{sec:res-corpus}. For the controlled LIBERO studies, evaluation comprises 2,000 episodes, 500 per suite. Our controlled LIBERO-Plus~\cite{plus} evaluations use a fixed manifest of 10,030 scenarios spanning seven perturbation categories. Baseline scores in \cref{sec:res-benchmarks} may use different protocols. The two benchmarks play complementary roles. Standard LIBERO evaluates the training distribution, on which strong policies exceed 95\%, whereas LIBERO-Plus perturbs object layouts, camera viewpoints, robot initial states, language instructions, lighting conditions, background textures, and sensor noise and discriminates between methods across the full range of scores reported here; the effects of interest are accordingly read primarily from LIBERO-Plus. Meta-World MT50~\cite{metaworld} evaluates a shared policy across 50 manipulation tasks. We use the episode-success criterion in the official evaluation protocol~\cite{metaworldeval}: an episode succeeds if the environment success flag is reached at any step. The reported MT50 score pools success over evaluation episodes.

\subsection{Training and Evaluation Protocol}
\label{sec:setup}
We use global batch size 320 and AdamW with $(\beta_1,\beta_2)=(0.9,0.95)$, weight decay 0.01, unit-norm gradient clipping, and no gradient accumulation. A cosine learning-rate schedule uses 5\% warmup and decays to 10\% of the peak rate. The released implementation assigns peak rates of $10^{-5}$ to the video generation model and policy blocks, and $10^{-4}$ to the geometry bridge, binding projections and gates, and non-block policy modules, including the action heads.

\cref{tab:protocol} summarizes the experimental groups. We separate the static-geometry training comparison, the mixed-data training routes, and the schedule--operator interaction. The branch-initialization study varies video-model and policy initialization independently. Dataset and static-training comparisons use static geometry; branch-initialization, noise, routing, learning-curve, and budget-sweep runs use multiplicative conditioning. Geometry-operator ablations use a reference initialization and a small-corpus robot-pretrained initialization, each compared with its own static-geometry baseline. Initialization probes use eight fixed LIBERO batches, with identical inputs, masks, positional encodings, and sampled noise across conditions.

\begin{table}[!htbp]
\centering
\caption{\textbf{Overview of experimental protocols.} Standard endpoint controls use 30k total downstream updates; target-domain S1 occupies the first 3k in the staged schedule. Learning curves also compare matched S2 updates, budget sweeps vary total updates, and extra pretraining is accounted for separately.}
\label{tab:protocol}
\setlength{\tabcolsep}{4pt}\small
\begin{tabular}{@{}P{0.23\linewidth}P{0.21\linewidth}P{0.18\linewidth}P{0.20\linewidth}P{0.10\linewidth}@{}}
\toprule
\textbf{Question} & \textbf{Intervention} & \textbf{Starting point} & \textbf{Reporting unit} & \textbf{Section}\\
\midrule
How far does the system scale? & Pretraining duration & Robot-data pretraining & Single-seed downstream score & \cref{sec:res-pretrain}\\
Does staging help? & S1--S2 vs. S2; static geometry & Pretrained Wan & Three-seed mean; paired SD & \cref{sec:res-static}\\
Which data route? & Target and mixed-data stages & Route-specific & Three-seed mean; paired SD & \cref{sec:res-corpus}\\
Does staging help across operators? & Schedule $\times$ operator & Pretrained Wan & Three-seed mean; paired SD & \cref{sec:factorial}\\
Which branch benefits? & $(W_0,W_1)\times(R_0,R_1)$ & Original / S1 weights & Single-run score & \cref{sec:transfer}\\
Which training choices matter? & Visual noise; gradient isolation & Original / S1 weights & Single-run score & \cref{sec:res-noise}\\
Which conditioning operator? & Operator variants; fixed background & Reference; robot-pretrained & Three-seed mean & \cref{sec:res-interface}\\
What changes at initialization? & Branch transfer; camera isolation & Four $(W_i,R_j)$ combinations & Eight-batch loss and gradients & \cref{sec:probes}\\
\bottomrule
\end{tabular}
\end{table}

\subsection{Evaluation Metrics and Statistical Reporting}
\label{sec:metrics}
For a benchmark with $N$ evaluated episodes, success rate is
\begin{equation}
\mathrm{SR}=\frac{100}{N}\sum_{e=1}^{N}\mathbf{1}[\text{episode }e\text{ succeeds}].
\label{eq:success-rate}
\end{equation}
The controlled LIBERO evaluations use 2,000 episodes and the controlled LIBERO-Plus evaluations use 10,030 scenarios. The complete-system, pretraining-sweep, and MT50 results are reported as aggregate success rates; their per-episode counts and repeat-evaluation details are not reported here. Success differences are absolute percentage points (pp), with the baseline stated for each comparison.

The staging, dataset, and geometry-operator studies report three-training-seed means. For paired seed-wise success differences $d_r$ between a variant and its specified baseline, we report the mean and sample standard deviation (SD):
\begin{equation}
\bar\Delta=\frac{1}{n}\sum_{r=1}^{n}d_r,\qquad
s_\Delta=\sqrt{\frac{1}{n-1}\sum_{r=1}^{n}(d_r-\bar\Delta)^2},\qquad n=3.
\label{eq:paired-sd}
\end{equation}
The notation $\bar\Delta\pm s_\Delta$ describes variability of the paired differences across training seeds, not the uncertainty of an individual model's success rate. Table difference columns give mean effects; SDs accompany the principal comparisons where reported. Geometry-operator comparisons with no reported SD are shown as mean differences only. Branch-initialization, noise, routing, learning-curve, budget-sweep, and pretraining results are single training runs. Fixed-batch probes and attention diagnostics are separate from training-seed statistics.

\subsection{Infrastructure and Inference}
\label{sec:infra}
All training runs use a single topology of four nodes with eight accelerators each and no gradient accumulation.

On one H800 at batch size one, mean decision latency is 331.25 ms and the 95th percentile is 365.30 ms. The measured planning rate is 3.02 plans per second. Executing four action steps per plan corresponds to an effective rate of 12.08 action steps per second. Total latency decomposes as
\begin{equation}
T_{\mathrm{infer}}=T_{\mathrm{enc}}+T_{\mathrm{world}}+N_{\mathrm{Euler}}T_{\mathrm{policy}}+T_{\mathrm{other}},
\label{eq:latency}
\end{equation}
where the terms denote encoding, one video-model forward pass, $N_{\mathrm{Euler}}=10$ policy evaluations, and other inference operations. The decomposition shows the effect of reading intermediate features instead of fully decoding frames. The video-model pass enters once per decision and is amortized across all ten solver steps and then across four executed actions, so the cost of the large module is incurred once per plan, and the term that scales with the number of solver steps is that of the ten-layer policy. The expression contains no frame-decoding term, because the action path reads features directly. The 95th-percentile latency exceeds the mean by 10.3\%.

\section{Main Results}
\label{sec:results}

The experiments ask whether the complete policy performs well in simulation and on a real robot, whether allocating a fixed target-task budget to S1 before S2 improves action learning, and which weight sources, noise supervision, and geometric coupling support that improvement. This section evaluates the complete system and the effect of additional robot-data pretraining. The controlled training comparisons follow in \cref{sec:staging}, branch/noise/geometry ablations in \cref{sec:ablations}, and initialization and attention diagnostics in \cref{sec:analysis}.

\subsection{Simulation Benchmarks}
\label{sec:res-benchmarks}
\begin{figure}[!htbp]
\centering
\paperfigure{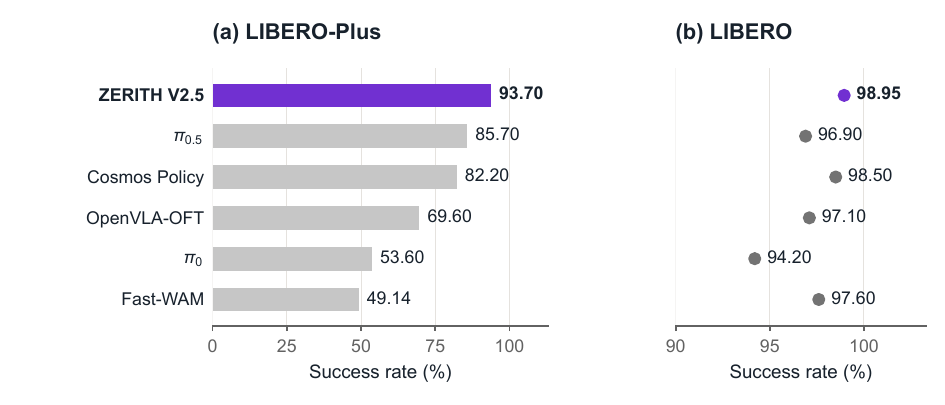}
\caption{\textbf{Reported benchmark scores.} The systems in \cref{tab:comparison}, ordered by LIBERO-Plus score in both panels. (a) LIBERO-Plus success; bars start at zero. (b) Standard LIBERO success; the point-plot axis starts at 90\%. Purple marks \method. Baseline protocols differ; both \method\ scores use one deployed checkpoint.}
\label{fig:reported-benchmarks}
\end{figure}

The deployed checkpoint achieves 98.95\% on standard LIBERO and 93.70\% on LIBERO-Plus (\cref{tab:comparison}); both scores come from a single checkpoint (\cref{fig:reported-benchmarks}). Among the listed baselines, Cosmos Policy has the highest standard LIBERO score, 98.5\%, and $\pi_{0.5}$ the highest LIBERO-Plus score, 85.7\%; \method\ exceeds them by 0.45 and 8.00 points, respectively. On LIBERO-Plus, the largest margin for \method{} among the listed baselines is 44.56 points over Fast-WAM. \cref{tab:comparison} also distinguishes video-model inference treatments. Cosmos Policy jointly predicts future states in its direct-policy mode but discards those predictions; \method\ reads future keys and values in a single video-model pass per decision without fully decoding a frame.

\begin{table}[!htbp]
\centering
\caption{\textbf{Benchmark success (\%).} Baseline scores are listed for comparison. Both bold scores correspond to one deployed checkpoint. K/V denotes attention keys and values; the final column describes video-model inference treatment. For Cosmos Policy, joint future-state prediction refers to the direct-policy mode used for its reported standard LIBERO score; the predictions are discarded.}
\label{tab:comparison}
\setlength{\tabcolsep}{5pt}\small
\begingroup
\fontsize{9.3}{12}\selectfont
\sisetup{group-digits=integer,detect-weight=true}
\begin{tabular*}{\linewidth}{@{\extracolsep{\fill}}llS[table-format=2.2]S[table-format=2.2]l@{}}
\toprule
\textbf{System} & \textbf{Backbone} & {\textbf{LIBERO} $\uparrow$} & {\textbf{Plus} $\uparrow$} & \textbf{Video inference}\\
\midrule
$\pi_0$~\cite{pi0} & PaliGemma 3B+0.3B & 94.20 & 53.60 & ---\\
$\pi_{0.5}$~\cite{pi05} & PaliGemma+flow & 96.90 & 85.70 & ---\\
OpenVLA-OFT~\cite{openvlaoft} & OpenVLA 7B & 97.10 & 69.60 & ---\\
Fast-WAM~\cite{fastwam} & Wan2.2-5B+1B & 97.60 & 49.14 & Current-frame K/V\\
Cosmos Policy~\cite{cosmospolicy} & Cosmos-Predict2 2B & 98.50 & 82.20 & \shortstack[l]{Joint future-state\\prediction}\\
\midrule
\method\ (Ours) & Wan2.2, 7B & \bfseries 98.95 & \bfseries 93.70 & Future-slot K/V\\
\bottomrule
\end{tabular*}
\endgroup

\end{table}

\paragraph{Additional multi-task evaluation.}
ZERITH V2.5 reaches 92.3\% on Meta-World MT50 (\cref{tab:mt50}), with success pooled over evaluation episodes. The reference scores, 47.9\% for $\pi_0$ as reported in SmolVLA~\cite{smolvla} and 81.6\% for $\pi_{0.5}$ as reported in LA4VLA~\cite{la4vla}, average four difficulty groups with equal weight. The 10.7-point difference from the listed $\pi_{0.5}$ score is therefore a cross-study comparison. This extends the complete-policy evaluation to a separate suite of 50 tasks.

\begin{table}[!htbp]
\centering
\caption{\textbf{Meta-World MT50 success (\%).} Reference results retain their source-specific evaluation protocols. The $\pi_0$ reference is the robot-pretrained variant in SmolVLA. We do not report whether the MT50 result uses the LIBERO deployment checkpoint.}
\label{tab:mt50}
\setlength{\tabcolsep}{5pt}\small
\begin{tabular}{@{}P{0.22\linewidth}S[table-format=2.1]P{0.58\linewidth}@{}}
\toprule
\textbf{System} & {\textbf{Success (\%)} $\uparrow$} & \textbf{Aggregation / source}\\
\midrule
$\pi_0$ & 47.9 & Four difficulty groups, equal weight; SmolVLA~\cite{smolvla}\\
$\pi_{0.5}$ & 81.6 & Four difficulty groups, equal weight; as reported in LA4VLA~\cite{la4vla}\\
\method & \bfseries 92.3 & Pooled evaluation episodes; reported complete-policy result\\
\bottomrule
\end{tabular}

\end{table}

These comparisons evaluate the combined policy, including additional robot-data pretraining and future-slot readout. Because architectures, pretraining, and evaluation protocols differ across systems, the benchmark margin cannot be attributed to S1 or to the readout architecture separately. The fixed-data, fixed-target-budget studies that follow test narrower design choices. \cref{sec:res-pretrain} varies additional robot-data pretraining while holding the downstream S1--S2 procedure fixed.

\subsection{Real-Robot Evaluation}
\label{sec:res-real}
\begin{figure}[!htbp]
\centering
\paperfigure{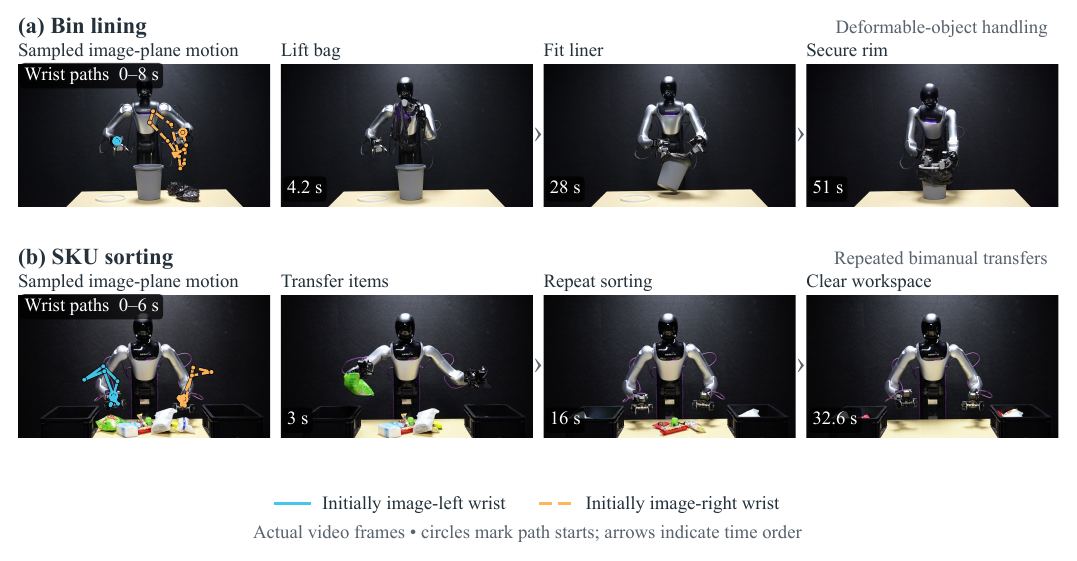}
\caption{\textbf{Real-robot task examples.} Each row contains a scene with wrist paths followed by three video keyframes. Paths connect manually annotated image-plane positions from the labeled intervals; colors identify wrists by their initial image-left or image-right positions. Frame labels give playback timestamps from the source recordings.}
\label{fig:task-trajectories}
\end{figure}

Rollout videos show deformable-object handling in bin lining and repeated bimanual transfers in SKU sorting on ZERITH H1. \cref{fig:task-trajectories} shows these two tasks, including the bin-lining and sorting sequences of \cref{fig:mechanism}, with timestamped frames and sampled image-plane wrist paths. The lining sequence shows the bag lifted, fitted around the bin, and secured at the rim; the sorting sequence shows the workspace progressively cleared into two side bins. The quantitative evaluation measures task-level success on pick and place and bin lining.

\cref{tab:real-tasks} reports task-level success for \method, $\pi_{0.5}$, and Fast-WAM, each evaluated over 100 pick-and-place trials and 20 bin-lining trials.

\begin{table}[!htbp]
\centering
\caption{\textbf{Task-level success (\%) on the real robot.} The second header row gives the number of trials per system.}
\label{tab:real-tasks}
\setlength{\tabcolsep}{5pt}\small
\begin{adjustbox}{max width=\linewidth}
\begin{tabular*}{0.80\linewidth}{@{\extracolsep{\fill}}lS[table-format=2.0]S[table-format=2.0]@{}}
\toprule
\textbf{System} & {\textbf{Pick and place} $\uparrow$} & {\textbf{Bin lining} $\uparrow$}\\
 & {100 trials} & {20 trials}\\
\midrule
Fast-WAM~\cite{fastwam} & 84 & 15\\
$\pi_{0.5}$~\cite{pi05} & 95 & 35\\
\method\ (Ours) & \bfseries 97 & \bfseries 40\\
\bottomrule
\end{tabular*}
\end{adjustbox}

\end{table}

\method\ attains the highest success on pick and place and bin lining. On pick and place it succeeds in 97 of 100 trials, compared with 95 for $\pi_{0.5}$ and 84 for Fast-WAM, that is, 3 failures against 5 and 16. On bin lining it succeeds in 8 of 20 trials (40\%), compared with 7 (35\%) for $\pi_{0.5}$ and 3 (15\%) for Fast-WAM. The margins are larger on bin lining, 5 points over $\pi_{0.5}$ and 25 over Fast-WAM, than on pick and place, 2 and 13 points, where \method\ and $\pi_{0.5}$ both succeed in at least 95 of 100 trials.

\subsection{Effect of Robot-Data Pretraining}
\label{sec:res-pretrain}
\begin{figure}[!htbp]
\centering
\paperfigure{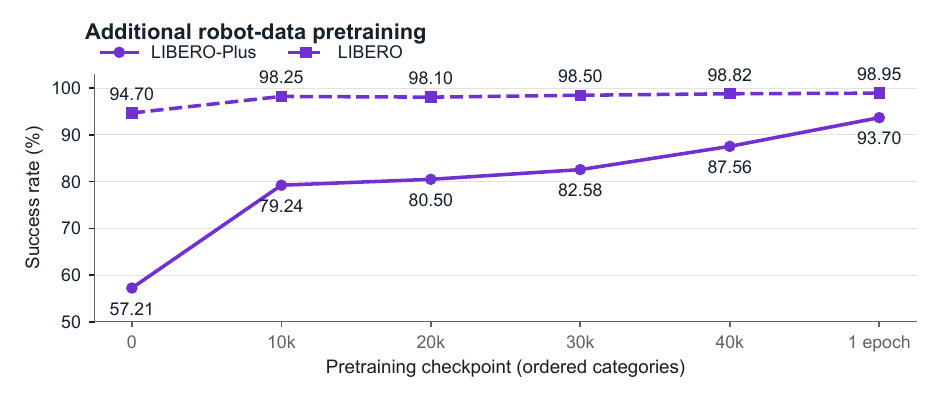}
\caption{\textbf{Performance after additional pretraining.} Aggregate LIBERO-Plus (solid circles) and standard LIBERO (dashed squares) success from single training runs. Both series evaluate the same downstream S1--S2 recipe for 30k updates. The zero-step reference omits robot-data pretraining. Checkpoints are ordered categories; aggregate values are listed in \cref{tab:pretraining}.}
\label{fig:pretraining}
\end{figure}

The benchmark scores of \cref{sec:res-benchmarks} are obtained after robot-data pretraining. To show how they depend on this stage, \cref{tab:pretraining} and \cref{fig:pretraining} compare a zero-robot-pretraining reference with five upstream checkpoints, each followed by an independent 30,000-update downstream adaptation using the same recipe. The zero-pretraining reference retains the original Wan initialization before downstream S1--S2 training. This experiment uses 10,000 hours of real-robot recordings selected after cleaning from public datasets and our ZERITH H1 platform described in \cref{sec:data}; the mixed-data ablation of \cref{sec:res-corpus} uses the smaller ABC-130K--LIBERO subset defined in \cref{sec:downstream}.

\begin{table}[!htbp]
\centering
\caption{\textbf{Effect of robot-data pretraining, aggregate success (\%), single training runs.} Every row uses 30k downstream S1--S2 updates. Zero steps omits robot-data pretraining while retaining the Wan initialization. Bold values identify the same deployed checkpoint.}
\label{tab:pretraining}
\setlength{\tabcolsep}{5pt}\small
\begin{tabular*}{0.80\linewidth}{@{\extracolsep{\fill}}lS[table-format=2.2]S[table-format=2.2]@{}}
\toprule
\textbf{Pretraining} & {\textbf{LIBERO (\%)} $\uparrow$} & {\textbf{LIBERO-Plus (\%)} $\uparrow$}\\
\midrule
0 steps & 94.70 & 57.21\\
10k & 98.25 & 79.24\\
20k & 98.10 & 80.50\\
30k & 98.50 & 82.58\\
40k & 98.82 & 87.56\\
1 epoch & \bfseries 98.95 & \bfseries 93.70\\
\bottomrule
\end{tabular*}

\end{table}

The two benchmarks differ substantially along this trajectory. Among the robot-pretrained checkpoints from 10,000 steps through one epoch, standard LIBERO ranges from 98.10\% to 98.95\%, a span of 0.85 points, while LIBERO-Plus rises monotonically from 79.24\% to 93.70\%, a gain of 14.46 points. The one-epoch checkpoint has the highest recorded score on both benchmarks. The near-saturated standard suite shows a smaller separation across the recorded checkpoints than the perturbation suite; We therefore read the remaining comparisons, like the controlled experiments, mainly from LIBERO-Plus. Relative to the configuration without robot-data pretraining (94.70\% and 57.21\%), the LIBERO-Plus gain is 22.03 points at 10,000 pretraining steps and 36.49 points after one epoch.

\section{Effect of Staged Training}
\label{sec:staging}

This section and the two that follow use the controlled setting of \cref{sec:setup}: standard endpoint controls share downstream demonstrations and 30,000 total updates, learning curves also compare matched S2 updates, and budget sweeps vary total updates. Additional pretraining is accounted for separately. This section measures how much staging changes success at a fixed budget (\cref{sec:res-static,sec:res-corpus,sec:factorial}) and how the gain evolves with the training budget (\cref{sec:res-crossover,sec:res-budget}); \cref{sec:ablations} isolates the components that produce it, and \cref{sec:analysis} characterizes the state that S1 creates. Static-training, data-route, and schedule--operator interaction results have separate tables, each labeled with its reporting unit.

\subsection{Staged Training with Static Geometry}
\label{sec:res-static}
We first ask whether allocating part of a fixed target-task budget to video-model adaptation improves the policy obtained from the remaining joint-training updates. Static geometry fixes the conditioning pathway, so only the training procedure changes. Both schedules use the same LIBERO demonstrations and 30,000 total updates: direct training uses 30,000 S2 updates, whereas staging uses 3,000 S1 updates followed by 27,000 S2 updates (\cref{tab:staging-static}).

\begin{table}[!htbp]
\centering
\caption{\textbf{S1--S2 versus direct S2 with static geometry.} Both schedules use the LIBERO target dataset and 30k total updates. Success rates are three-seed means (\%); $\Delta$Plus is the mean percentage-point difference from direct S2.}
\label{tab:staging-static}
\begingroup
\fontsize{9.3}{12}\selectfont
\sisetup{group-digits=integer}
\begin{tabular*}{\linewidth}{@{\extracolsep{\fill}}lS[table-format=2.2]S[table-format=2.2]S[table-format=+2.2]@{}}
\toprule
\textbf{Target training} & {\textbf{LIBERO} $\uparrow$} & {\textbf{Plus} $\uparrow$} & {$\boldsymbol{\Delta}$\textbf{Plus (pp)} $\uparrow$}\\
\midrule
S2 ($D_L$; 30k), baseline & 92.15 & 37.64 & +0.00\\
S1 ($D_L$; 3k) $\to$ S2 ($D_L$; 27k) & 95.30 & 56.80 & +19.16\\
\bottomrule
\end{tabular*}
\endgroup

\end{table}

S1--S2 raises LIBERO-Plus success from 37.64\% to 56.80\%. The paired gain is $19.16\pm0.26$ points (mean $\pm$ SD); on standard LIBERO it is $3.15\pm0.40$ points, from 92.15\% to 95.30\%. The three recorded Plus gains are 19.16, 19.41, and 18.90 points. Allocating part of the same target-task budget to video-model adaptation therefore improves control in this setting, especially under the tested perturbations. This evaluates the broad-noise S1--S2 procedure as a whole; \cref{sec:res-noise} tests S1 noise coverage separately.

\subsection{Effect of the Adaptation Dataset}
\label{sec:res-corpus}
The mixed-data comparison asks whether broader robot experience makes target-domain video-model adaptation unnecessary. Let $D_L$ denote the LIBERO target demonstrations and $D_M$ the mixture of 13,259 ABC-130K episodes and 1,693 LIBERO demonstrations. Starting from S1 on $D_M$, we compare direct target S2 with a route that inserts S1 on $D_L$ (\cref{tab:data-routes}). Every route uses 30,000 target-task updates: 3,000 S1 plus 27,000 S2 when target S1 is present, and 30,000 S2 otherwise. Stages on $D_M$ precede this target-task budget and count as additional pretraining.

\begin{table}[!htbp]
\centering
\caption{\textbf{Training routes with mixed data $D_M$ and target data $D_L$.} Arrows show stage order. Every route receives 30k target-task updates; upstream mixed-data stages are additional. Success rates are three-seed means (\%).}
\label{tab:data-routes}
\begingroup
\fontsize{9.3}{12}\selectfont
\sisetup{group-digits=integer}
\begin{tabular*}{\linewidth}{@{\extracolsep{\fill}}lS[table-format=2.2]S[table-format=2.2]@{}}
\toprule
\textbf{Training sequence} & {\textbf{LIBERO} $\uparrow$} & {\textbf{LIBERO-Plus} $\uparrow$}\\
\midrule
S1 ($D_L$) $\to$ S2 ($D_L$) & 95.30 & 56.80\\
S1 ($D_M$) $\to$ S2 ($D_L$) & 96.15 & 55.56\\
S1 ($D_M$) $\to$ S2 ($D_M$) $\to$ S2 ($D_L$) & 96.75 & 63.85\\
S1 ($D_M$) $\to$ S1 ($D_L$) $\to$ S2 ($D_L$) & 95.92 & 59.73\\
\bottomrule
\end{tabular*}
\endgroup

\end{table}

\begin{figure}[!htbp]
\centering
\paperfigure{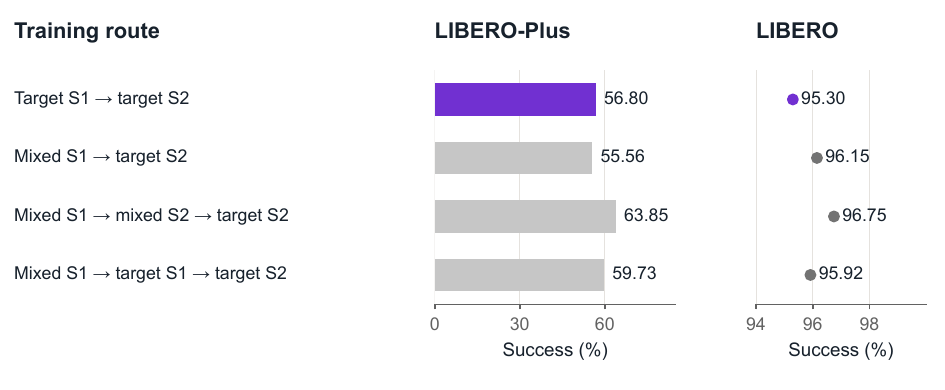}
\caption{\textbf{Dataset and training-stage ablations.} Four routes under static-only geometry, three-seed means. Every route ends with target-task S2; mixed denotes ABC robot data plus LIBERO and target denotes LIBERO. Purple identifies the target-S1 recipe. Plus bars start at zero; the standard-LIBERO point axis starts at 94\%.}
\label{fig:corpus-controls}
\end{figure}

Adding target-domain S1 after mixed-data S1 raises LIBERO-Plus success from 55.56\% to 59.73\%, a gain of 4.17 points. A separate route adding mixed-data S2 reaches 63.85\%, 8.29 points above mixed-data S1 followed directly by target S2. This shows that target-domain adaptation remains useful after mixed-data video-model adaptation, while action-supervised robot pretraining can also improve the target-task starting point. The latter route adds an upstream stage, so the 4.12-point difference between the two extended routes also reflects extra compute. \cref{fig:corpus-controls} shows the complete sequences.

Target-only S1 reaches 56.80\% LIBERO-Plus success versus 55.56\% for mixed-data S1 while using only 11.3\% as many episodes; the paired difference is $1.23\pm0.14$ points (mean $\pm$ SD). On standard LIBERO the ordering is reversed: mixed-data S1 reaches 96.15\% versus 95.30\%, a difference of 0.85 points. Target-only S1 ranks higher on the perturbation suite, and mixed-data S1 on the standard suite.

\subsection{Staging Across Geometric Operators}
\label{sec:factorial}
This study measures the effect of changing the training schedule under each geometric-conditioning operator. The matched $2\times3$ comparison crosses direct S2 and S1--S2 with static, additive, and multiplicative conditioning (\cref{tab:factorial}). All runs use the same LIBERO demonstrations and 30,000 total target-task updates. This interaction test is distinct from the geometry-operator ablations in \cref{sec:res-interface}, which hold initialization and training fixed within each background.

\begin{table}[!htbp]
\centering
\caption{\textbf{Interaction between training schedule and geometric conditioning.} Success rates are three-seed means (\%). $\Delta$Plus is the mean difference in pp from static geometry within the same schedule. All runs use the LIBERO target dataset and 30k total updates.}
\label{tab:factorial}
\begingroup
\fontsize{9.3}{12}\selectfont
\sisetup{group-digits=integer}
\begin{tabular*}{\linewidth}{@{\extracolsep{\fill}}llS[table-format=2.2]S[table-format=2.2]S[table-format=+1.2]@{}}
\toprule
\textbf{Target training} & {\boldmath$F(g,d)$} & {\textbf{LIBERO} $\uparrow$} & {\textbf{Plus} $\uparrow$} & {$\boldsymbol{\Delta}$\textbf{Plus (pp)} $\uparrow$}\\
\midrule
S2 & $0$ (baseline) & 92.15 & 37.64 & +0.00\\
S2 & $g+d$ & 91.70 & 38.70 & +1.06\\
S2 & $g\odot d$ & 91.80 & 39.93 & +2.29\\
\midrule
S1 $\to$ S2 & $0$ (baseline) & 95.30 & 56.80 & +0.00\\
S1 $\to$ S2 & $g+d$ & 95.55 & 56.63 & -0.17\\
S1 $\to$ S2 & $g\odot d$ & 95.65 & 57.48 & +0.68\\
\bottomrule
\end{tabular*}
\endgroup

\end{table}

\begin{figure}[!htbp]
\centering
\paperfigure{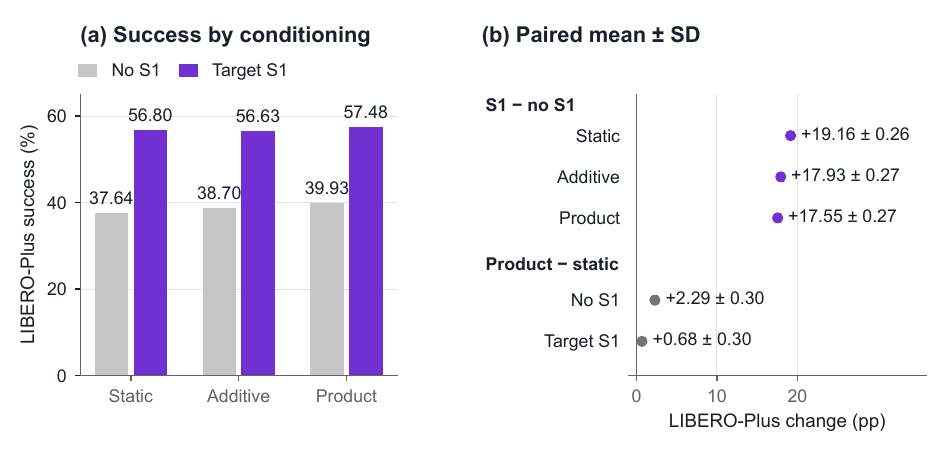}
\caption{\textbf{Schedule--operator interaction.} (a) Three-seed mean LIBERO-Plus success for the six combinations. (b) Paired mean differences with error bars of $\pm1$ sample SD across three training seeds. The upper group compares S1--S2 with direct S2 within each operator; the lower group compares product with static conditioning within each schedule. Differences are in pp.}
\label{fig:factorial}
\end{figure}

The S1--S2 gains on LIBERO-Plus are $19.16\pm0.26$, $17.93\pm0.27$, and $17.55\pm0.27$ points under static, additive, and multiplicative conditioning, respectively (mean $\pm$ paired SD). Product-over-static gains are $2.29\pm0.30$ points under direct S2 and $0.68\pm0.30$ points under S1--S2. Thus the staging benefit persists under every tested operator. The improvement on standard LIBERO is 3.15 points with static geometry and 3.85 points with either of the other operators; the Plus gains are 4.6--6.1 times these standard-suite gains. These are ratios of mean effects (\cref{fig:factorial}).

\FloatBarrier

The product-over-static gain decreases from 2.29 points without S1 to 0.68 points with S1, an interaction point estimate of $-1.61$ points, equal to $17.55-19.16$. This attenuation is consistent with partially overlapping contributions from video-model adaptation and geometric coupling. The attention shift toward adapted-video-model features in \cref{sec:res-attention} points the same way, although it comes from a different intervention and does not by itself confirm the explanation.

\subsection{Learning Curves and Update Efficiency}
\label{sec:res-crossover}
\begin{figure}[!htbp]
\centering
\paperfigure{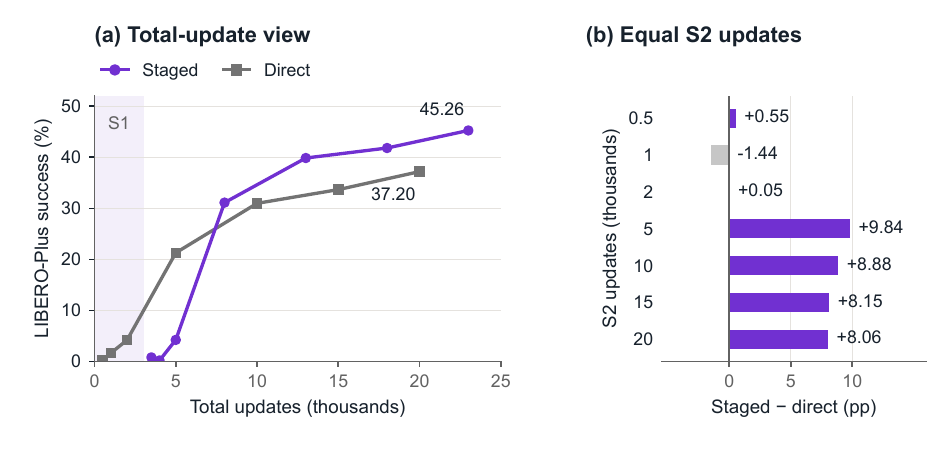}
\caption{\textbf{LIBERO-Plus learning checkpoints.} (a) The recorded learning checkpoints against total updates, including 3k S1 updates for staged training. Lines connect recorded points as visual guides. (b) Registered staged-minus-direct differences at equal S2 updates; these comparisons give staged training 3k more total updates. Budget-sweep endpoints are shown separately in \cref{sec:res-budget}; exact values and unequal-budget comparisons remain in \cref{tab:early,tab:matched}.}
\label{fig:crossover-observed}
\end{figure}

A higher final score leaves open whether S1 repays the early updates it takes away from action learning. We therefore compare policy checkpoints against both total updates, including S1, and S2 updates. With LIBERO demonstrations and multiplicative conditioning fixed, direct S2 starts from $(W_0,R_0)$ and the staged curve uses $(W_1,R_1)$. The learning-curve model of \cref{sec:discussion} makes a specific prediction about the early part of training: under its constant-rate assumptions, the staged schedule first falls below direct training and then overtakes it once the faster second stage has accumulated sufficient additional progress. \cref{tab:early} tests this prediction by evaluating both schedules at seven checkpoints from 500 to 20,000 S2 updates. Two budget conventions apply to the same checkpoints and answer different questions. Matching S2 updates compares the schedules after equal numbers of joint updates, although the routing schedule is defined relative to each run's S2 length (\cref{sec:practice}); matching total updates measures which schedule achieves higher success per update, and is the convention under which the crossover is defined (\cref{fig:crossover-observed}).

\begin{table}[!htbp]
\centering
\caption{\textbf{Budget curve on LIBERO-Plus.} ``Total'' is all updates consumed, so the staged column includes its 3,000 S1 updates. Percentages are over the 10,030-scenario manifest. The last two rows give the 30k-total endpoints. Here $\Delta$ is staged minus direct at matched S2 updates, in pp.}
\label{tab:early}
\setlength{\tabcolsep}{5pt}\small
\begingroup
\fontsize{9.3}{12}\selectfont
\renewcommand{\arraystretch}{1.08}
\begin{tabular*}{\linewidth}{@{\extracolsep{\fill}}S[table-format=5.0]S[table-format=5.0]S[table-format=2.2]S[table-format=5.0]S[table-format=2.2]S[table-format=+1.2]@{}}
\toprule
{} & \multicolumn{2}{c}{\textbf{Direct joint}} & \multicolumn{2}{c}{\textbf{Staged (3k S1)}} & {}\\
\cmidrule(lr){2-3}\cmidrule(lr){4-5}
{\textbf{S2 updates}} & {\textbf{Total}} & {\textbf{Plus} $\uparrow$} & {\textbf{Total}} & {\textbf{Plus} $\uparrow$} & {$\boldsymbol{\Delta}\uparrow$}\\
\midrule
500 & 500 & 0.24 & 3,500 & 0.79 & +0.55\\
1,000 & 1,000 & 1.63 & 4,000 & 0.19 & -1.44\\
2,000 & 2,000 & 4.17 & 5,000 & 4.22 & +0.05\\
5,000 & 5,000 & 21.28 & 8,000 & 31.12 & +9.84\\
10,000 & 10,000 & 30.98 & 13,000 & 39.86 & +8.88\\
15,000 & 15,000 & 33.67 & 18,000 & 41.82 & +8.15\\
20,000 & 20,000 & 37.20 & 23,000 & 45.26 & +8.06\\
\midrule
27,000 & {---} & {---} & 30,000 & 56.70 & {---}\\
30,000 & 30,000 & 37.54 & {---} & {---} & {---}\\
\bottomrule
\end{tabular*}
\endgroup

\end{table}

At matched S2 updates, the staged-minus-direct differences are $+0.55$, $-1.44$, and $+0.05$ points at 500, 1,000, and 2,000 updates, during which both schedules remain below 5\% success. From 5,000 S2 updates onward the differences are $+9.84$, $+8.88$, $+8.15$, and $+8.06$ points at 5,000, 10,000, 15,000, and 20,000 updates. After the first few thousand policy updates, the policy attached to the adapted video generation model thus achieves 8 to 10 points higher success than the policy attached to the original video generation model at every recorded checkpoint.

On the total-update axis the recorded point estimates indicate an early crossover. At 5,000 total updates, staged training reaches 4.22\% after 3,000 S1 and 2,000 joint updates, versus 21.28\% for direct training, a 17.06-point deficit. The staged checkpoints at 3,500 and 4,000 updates score 0.79\% and 0.19\%; no direct checkpoints at these exact budgets are reported. At 8,000 total updates, the staged checkpoint reaches 31.12\%, slightly above the direct checkpoint at 10,000 updates (30.98\%). A 5,000--8,000-update interval ($1.7\tau$--$2.7\tau$ for $\tau=3{,}000$) is exploratory: using 8,000 as an upper bound assumes that the unmeasured direct score there does not exceed its 10,000-update score. The 0.14-point difference is within sampling error. The broader 5,000--13,000-update interval likewise relies on unmeasured equal-budget points (\cref{sec:limitations}). For the illustrative linear model, we assume $r_{AC}=0$ for a hypothetical common control probe during S1, which we do not measure. Conditional on the narrower interval, \cref{eq:inversion} gives a rate ratio of 1.6--2.5; this value follows from the model. All candidate crossings lie after S1, consistent with $T^*>\tau$ in \cref{eq:crossover}.

\begin{table}[!htbp]
\centering
\caption{\textbf{Checkpoint comparisons at matched or exceeded success levels.} For every recorded direct checkpoint, the first recorded staged checkpoint with higher LIBERO-Plus success. ``Staged budget'' is staged updates as a percentage of direct updates; exposures saved are the update difference times global batch 320. Values below 100\% indicate fewer updates relative to the listed direct checkpoint.}
\label{tab:matched}
\setlength{\tabcolsep}{5pt}\small
\begingroup
\fontsize{9.3}{12}\selectfont
\renewcommand{\arraystretch}{1.08}
\begin{tabular*}{\linewidth}{@{\extracolsep{\fill}}S[table-format=5.0]S[table-format=2.2]S[table-format=5.0]S[table-format=2.2]S[table-format=3.1]S[table-format=+2.2]@{}}
\toprule
\multicolumn{2}{c}{\textbf{Direct checkpoint}} & \multicolumn{2}{c}{\shortstack{\textbf{First staged checkpoint}\\\textbf{above it}}} & {\textbf{Staged budget} $\downarrow$} & {\textbf{Exposures} $\uparrow$}\\
\cmidrule(lr){1-2}\cmidrule(lr){3-4}
{\textbf{Updates}} & {\textbf{Plus} $\uparrow$} & {\textbf{Updates}} & {\textbf{Plus} $\uparrow$} & {(\% of direct)} & {saved (M)}\\
\midrule
500 & 0.24 & 3,500 & 0.79 & 700.0 & -0.96\\
1,000 & 1.63 & 5,000 & 4.22 & 500.0 & -1.28\\
2,000 & 4.17 & 5,000 & 4.22 & 250.0 & -0.96\\
5,000 & 21.28 & 8,000 & 31.12 & 160.0 & -0.96\\
\midrule
10,000 & 30.98 & 8,000 & 31.12 & 80.0 & 0.64\\
15,000 & 33.67 & 13,000 & 39.86 & 86.7 & 0.64\\
20,000 & 37.20 & 13,000 & 39.86 & 65.0 & 2.24\\
30,000 & 37.54 & 13,000 & 39.86 & 43.3 & 5.44\\
35,000 & 38.46 & 13,000 & 39.86 & 37.1 & 7.04\\
40,000 & 37.88 & 13,000 & 39.86 & 32.5 & 8.64\\
50,000 & 38.52 & 13,000 & 39.86 & 26.0 & 11.84\\
\bottomrule
\end{tabular*}
\endgroup

\end{table}

At the later recorded checkpoints, staged training reaches the referenced success levels with fewer updates. \cref{tab:matched} pairs each direct checkpoint with the first staged checkpoint that exceeds it. For the four earliest direct reference checkpoints, the staged schedule requires more updates, between 1.6 and 7.0 times the direct budget, as the model predicts. From the 10,000-update direct reference onward, every listed staged checkpoint uses fewer updates. Savings are 20.0\% and 13.3\% relative to the 10,000- and 15,000-update direct references, and 74.0\% relative to the 50,000-update reference; they rise and fall across rows. A single staged checkpoint has a higher point estimate than the remaining evaluated direct checkpoints: at 13,000 updates the staged schedule reaches 39.86\%, above every evaluated direct checkpoint up to 50,000 updates, the highest of which is 38.52\%. The resulting 1.34-point margin over the 50,000-update score is within the sampling error of these single-run evaluations (\cref{sec:limitations}). Against the direct 30,000-update endpoint of 37.54\%, staging uses 43.3\% of the updates and saves 17,000 updates, or 5.44M sample exposures at global batch 320. Against the 50,000-update direct checkpoint, it uses 26.0\% of the updates and saves 37,000 updates, or 11.84M sample exposures. Because both schedules draw from the same 1,693 demonstrations, these exposures are repeated passes over a fixed dataset; the saving is in training passes over that data.

The recorded staged checkpoints at 23,000 and 30,000 total updates score 45.26\% and 56.70\%, a difference of 11.44 points over 7,000 updates. The direct checkpoints at 20,000 and 30,000 updates score 37.20\% and 37.54\%, a difference of 0.34 points. The 19.16-point staged-minus-direct difference is measured at matched 30,000 total updates; the earlier 8--10-point differences use matched S2 updates.

In these comparisons, staging is more efficient in two respects. For the later recorded checkpoints in \cref{tab:matched}, it reaches the referenced success levels with fewer updates. It also processes fewer input tokens during the 3,000 S1 updates, which omit the policy and geometry pathways; at a matched 30,000-update budget this omits 10\% of all action-token exposures (\cref{eq:tokens}). \cref{sec:theory-cost} combines update counts with assumed per-update costs.

\subsection{Performance Across Total Update Budgets}
\label{sec:res-budget}
The learning curves probe when the staged schedule overtakes direct training. A complementary question is whether the advantage persists when direct training is granted a larger budget. \cref{tab:budget} compares both schedules at five matched total budgets from 20,000 to 50,000 updates, each evaluated as an independent endpoint. Standard LIBERO is reported at all five budgets and LIBERO-Plus at all direct budgets and at the four staged budgets from 30,000 to 50,000 updates. The 30,000-update row reports the endpoint of the budget sweep.

\begin{table}[!htbp]
\centering
\caption{\textbf{Budget sweep.} Success (\%) at five total update budgets; the staged schedule allocates its first 3,000 updates to S1. Standard denominators are 2,000 episodes and Plus denominators 10,030 scenarios. Bold indicates the highest success rate in each column; $\Delta$ is staged minus direct in pp. A dash marks a budget that was not evaluated.}
\label{tab:budget}
\setlength{\tabcolsep}{5pt}\small
\begin{adjustbox}{max width=\linewidth}
\begin{tabular*}{\linewidth}{@{\extracolsep{\fill}}lS[table-format=2.2]S[table-format=2.2]S[table-format=+1.2]S[table-format=2.2]S[table-format=2.2]S[table-format=+2.2]@{}}
\toprule
 & \multicolumn{3}{c}{\textbf{Standard LIBERO} $\uparrow$} & \multicolumn{3}{c}{\textbf{LIBERO-Plus} $\uparrow$}\\
\cmidrule(lr){2-4}\cmidrule(lr){5-7}
\textbf{Budget} & {\textbf{Direct}} & {\textbf{Staged}} & {$\boldsymbol{\Delta}\uparrow$} & {\textbf{Direct}} & {\textbf{Staged}} & {$\boldsymbol{\Delta}\uparrow$}\\
\midrule
20k & 88.70 & 91.80 & +3.10 & 37.20 & {---} & {---}\\
30k & 92.35 & 95.35 & +3.00 & 37.54 & 56.70 & +19.16\\
35k & 92.95 & \bfseries 96.05 & +3.10 & 38.46 & \bfseries 57.08 & +18.61\\
40k & \bfseries 93.15 & 94.80 & +1.65 & 37.88 & 54.97 & +17.09\\
50k & 92.05 & 94.55 & +2.50 & \bfseries 38.52 & 55.20 & +16.68\\
\bottomrule
\end{tabular*}
\end{adjustbox}

\end{table}

\begin{figure}[!htbp]
\centering
\paperfigure{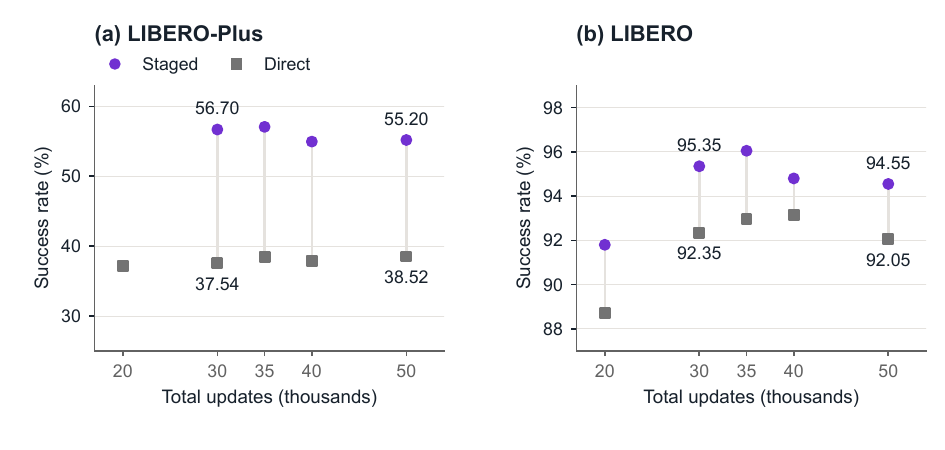}
\caption{\textbf{Success at reported total training budgets.} Purple circles denote staged and gray squares direct training. Vertical segments pair the two schedules' endpoints at the same budget. Staged LIBERO-Plus at 20k is unreported. Exact scores and registered differences are retained in \cref{tab:budget}.}
\label{fig:budget}
\end{figure}

Staged training achieves higher success than direct training at every budget for which both schedules are evaluated, on both benchmarks (\cref{fig:budget}). Because the staged total at each budget includes its 3,000 S1 updates, these endpoints show that the adaptation cost is recovered in update count at every evaluated LIBERO-Plus budget from 30,000 updates onward. The staged schedule at 30,000 updates exceeds direct training at 50,000 by 18.18 points on LIBERO-Plus, 56.70\% against 38.52\%, with 40\% fewer updates. Across the five budgets, direct training varies between 37.20\% and 38.52\%, a range of 1.32 points. The staged schedule reaches its highest LIBERO-Plus score, 57.08\%, at 35,000 updates, and records 54.97\% and 55.20\% at 40,000 and 50,000 updates. Both schedules rise and fall across the sweep.

The LIBERO-Plus advantage is 19.16, 18.61, 17.09, and 16.68 points at 30,000, 35,000, 40,000, and 50,000 updates, a reduction of 2.48 points over 20,000 additional updates, or 1.24 points per 10,000. This is an average over the interval; \cref{eq:area} relates it to the continuous learning curves. A margin of 16.68 points remains at 50,000 updates. On standard LIBERO, where scores range from 88.70\% to 96.05\% across both schedules, the staged advantage ranges from 1.65 to 3.10 points. At the 30,000-update budget, the effect of video-model adaptation is 19.16 points on LIBERO-Plus and 3.00 points on standard LIBERO, a ratio of 6.4.

\section{Ablation Studies}
\label{sec:ablations}

This section reports controlled variations at the 30,000-update budget: the initialization of each branch (\cref{sec:transfer}), S1 noise coverage and action-to-video-model gradient routing (\cref{sec:res-noise}), and the geometric conditioning and policy interface (\cref{sec:res-interface}).

\subsection{Video-Model and Policy Initialization}
\label{sec:transfer}
The static-training and interaction studies fix the policy initialization at $R_0$. The branch-transfer experiment asks whether S1 benefits control through the video-model representation, the policy initialization, or both. \cref{tab:controls} examines branch transfer by crossing the two weight sources: the video generation model is initialized from original Wan weights ($W_0$) or from S1-adapted weights ($W_1$), and the policy independently from the selected layers of either ($R_0$ or $R_1$). Direct training uses 30,000 S2 updates; each transfer configuration uses 27,000 S2 updates after the 3,000-update S1 run that supplies its adapted weights, including $(W_0,R_1)$, whose video generation model is restored to $W_0$. The direct $(W_0,R_0)$ score of 37.54\% is a single-run result; the corresponding multiplicative cell of the three-seed factorial study (\cref{tab:factorial}) reports a mean of 39.93\%. Relative to direct training with $(W_0,R_0)$, transferring only the adapted video-model weights, $(W_1,R_0)$, improves LIBERO-Plus by 19.67 points and standard LIBERO by 2.35 points. Transferring only the adapted policy weights, $(W_0,R_1)$, changes the two scores by $-3.94$ and $-3.55$ points and gives the lowest success of the four combinations on both benchmarks. In these comparisons, the gain of staging follows the adapted video-model weights. The fixed-batch loss and gradient probes in \cref{sec:probes} measure the starting state of these four initializations before S2.

\begin{table}[!htbp]
\centering
\caption{\textbf{Branch-initialization ablations.} Single-run success rates (\%) at 30k total updates with multiplicative conditioning. $R_1$ copies selected S1 video-model layers.}
\label{tab:controls}
\setlength{\tabcolsep}{5pt}\small
\begingroup
\fontsize{9.3}{12}\selectfont
\sisetup{group-digits=integer,detect-weight=true}
\begin{tabular*}{\linewidth}{@{\extracolsep{\fill}}lS[table-format=2.2]S[table-format=2.2]@{}}
\toprule
\textbf{Training / initialization} & {\textbf{LIBERO (\%)} $\uparrow$} & {\textbf{Plus (\%)} $\uparrow$}\\
\midrule
\multicolumn{3}{@{}l}{\textit{Branch initialization}}\\
\addlinespace[2pt]
Direct joint: $(W_0,R_0)$ & 92.35 & 37.54\\
Staged, video-model transfer: $(W_1,R_0)$ & 94.70 & \bfseries 57.21\\
Policy transfer only: $(W_0,R_1)$ & 88.80 & 33.60\\
Staged, both transferred: $(W_1,R_1)$ & \bfseries 95.35 & 56.70\\
\bottomrule
\end{tabular*}
\endgroup

\end{table}

In the policy-only configuration, the policy is copied from a checkpoint that is better on every recorded visual-objective measure. The same adapted weights thus increase success when placed in the video generation model and decrease it when placed in the policy. We interpret this as functional specialization: S1 trains visual prediction, whereas the policy extracts action information from a representation. The local analysis in \cref{sec:theory-transfer} offers a possible account in which the effect of a weight displacement depends on the receiving objective's gradient and curvature. We do not measure these quantities, so this account remains a hypothesis.

Once the video generation model is adapted, the choice of policy initialization has a much smaller effect. With $W_1$ fixed, $R_0$ scores 0.51 points higher on LIBERO-Plus, whereas $R_1$ scores 0.65 points higher on standard LIBERO, so the two benchmarks rank the policy initializations differently. Both configurations exceed direct training by more than 19 points on LIBERO-Plus, and the video-model-transfer effect is 39 times the difference between the two policy initializations. We deploy $(W_1,R_0)$, which attains the highest LIBERO-Plus score of the four.

\subsection{Noise Schedule and Gradient Routing}
\label{sec:res-noise}
\begin{figure}[!htbp]
\centering
\paperfigure{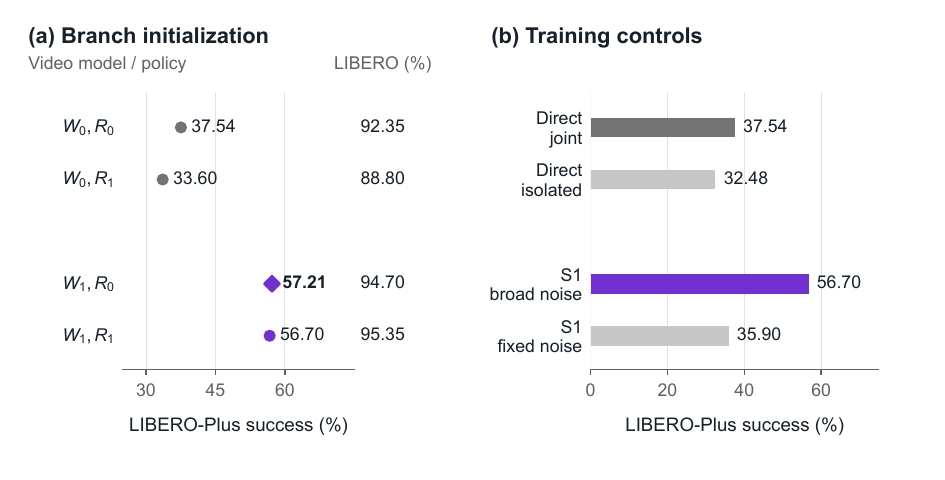}
\caption{\textbf{Branch initialization and training controls.} Single-run success at 30k total updates. (a) LIBERO-Plus point estimates for the four $(W_i,R_j)$ combinations; the adjacent numeric column gives standard LIBERO. Purple identifies adapted-video-model initialization, and the diamond marks the deployed $(W_1,R_0)$ configuration. (b) Direct joint training, direct training with action-to-video-model gradients blocked throughout, staged broad-noise S1, and fixed-noise S1. Broad-noise S1 uses the $(W_1,R_1)$ result for the noise comparison. Both benchmarks are retained in \cref{tab:controls,tab:noise-controls}.}
\label{fig:transfer-controls}
\end{figure}

S1 differs from direct joint training in two respects at once. It trains the video generation model on the visual objective alone, and it covers the full noise interval, whereas S2 reads the backbone at a single noise level. We evaluate two complementary controls: video-model-only S1 restricted to the pure-noise endpoint, and direct training with action-to-video-model gradients blocked throughout. \cref{tab:noise-controls} reports their downstream scores (\cref{fig:transfer-controls}).

\begin{table}[!htbp]
\centering
\caption{\textbf{S1 noise and S2 gradient-routing controls.} All runs use LIBERO target data, 30k total updates, and multiplicative conditioning. Success rates (\%) are single-run results. Both S1 variants copy the policy from their respective S1 video-model layers.}
\label{tab:noise-controls}
\begingroup
\fontsize{9.3}{12}\selectfont
\sisetup{group-digits=integer}
\begin{tabular*}{\linewidth}{@{\extracolsep{\fill}}llS[table-format=2.2]S[table-format=2.2]@{}}
\toprule
\textbf{S1 noise schedule} & \textbf{S2 action-to-video-model gradient} & {\textbf{LIBERO} $\uparrow$} & {\textbf{Plus} $\uparrow$}\\
\midrule
No S1 (baseline) & Blocked $\to$ ramped $\to$ full & 92.35 & 37.54\\
Broad noise & Blocked $\to$ ramped $\to$ full & 95.35 & 56.70\\
Fixed $\sigma_w=1$ & Blocked $\to$ ramped $\to$ full & 91.10 & 35.90\\
No S1 & Blocked throughout ($\beta=0$) & 90.90 & 32.48\\
\bottomrule
\end{tabular*}
\endgroup

\end{table}

The first ablation restricts S1 to $\sigma_w=1$, so that adaptation takes place at exactly the operating point that the policy later uses. Both noise conditions transfer $(W_1,R_1)$ from their respective S1 runs: each retains its adapted video-model weights and copies the selected S1 video-model layers into the policy. Only the initialization rule is shared between the two conditions. Fixed-noise S1 reaches 35.90\% LIBERO-Plus success, 20.80 points below broad-noise S1 and 1.64 points below direct training, and its standard LIBERO score is 4.25 points below broad-noise S1. The difference between the two noise schedules exceeds the entire staging effect. These single-run scores point to a benefit from broader noise coverage, although the two runs may also differ in other respects. \cref{sec:theory-noise} gives a population-level account of the additional structure that the interior of the noise interval identifies.

The second ablation retains direct training but removes the action-loss gradient to the video generation model for the whole run, so that the video generation model is updated only by its visual loss, as in S1, while the policy is trained concurrently. This configuration reaches 32.48\% LIBERO-Plus success, 5.06 points below direct training, and lowers standard success by 1.45 points. This single-run result supports the action objective as a useful training signal for the video-model representation once the policy is attached; staging postpones this signal until the video generation model has been adapted and then restores it.

Both controls score below standard direct training in these single-run comparisons. They support a role for both broad-noise adaptation and action-gradient feedback in the evaluated configurations. Because the isolation control has no S1, it leaves open whether permanent isolation after S1 would also hurt, and how long the gradient ramp should be.

\subsection{Geometry and Policy-Interface Ablations}
\label{sec:res-interface}
\begin{figure}[!htbp]
\centering
\paperfigure{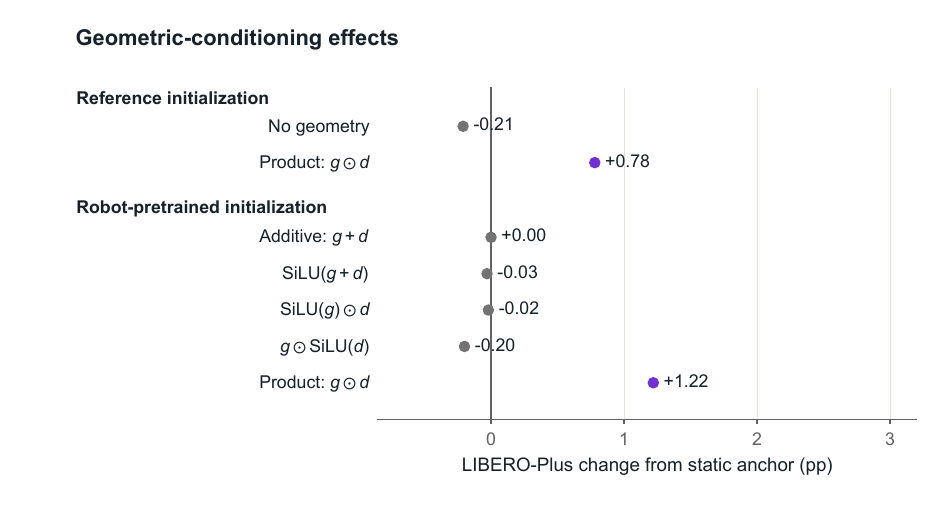}
\caption{\textbf{Geometric-conditioning operator effects.} Mean LIBERO-Plus changes from the static-geometry baseline within each initialization setting. Purple identifies the bilinear product. Points show mean differences only; no error bars are reported for these comparisons. $g$ is projected static geometry and $d$ is projected video-feature change.}
\label{fig:geometry}
\end{figure}

This study fixes the training procedure within each initialization background and varies the geometry operator or policy interface. The schedule--operator interaction is evaluated separately in \cref{sec:factorial}. We use two fixed checkpoint settings: reference initialization and small-corpus robot-pretrained initialization. Within each setting, all variants use the same initial checkpoint, LIBERO target data, training procedure, and 30,000 downstream updates, and are compared with that setting's static-geometry baseline (\cref{tab:geometry,fig:geometry}). Static geometry retains memory $(K_g,V_g)$ while setting the residual $F(g,d)=0$. Other rows change $F$; the no-geometry row also removes the geometry memory. This separates the presence of geometry information from the form of its coupling. The product improves LIBERO-Plus by 0.78 points in the reference setting and 1.22 points in the robot-pretrained setting. Removing geometry in the reference setting changes success by $-0.21$ points; this difference is small, and the three-seed mean has no reported SD.

\begin{table}[!htbp]
\centering
\caption{\textbf{Geometric-conditioning operator ablations.} Success rates are three-seed means (\%). $\Delta$Plus is the mean difference in pp from the marked static-geometry baseline within the same initialization block.}
\label{tab:geometry}
\begingroup
\fontsize{9.3}{12}\selectfont
\sisetup{group-digits=integer}
\begin{tabular*}{\linewidth}{@{\extracolsep{\fill}}llS[table-format=2.2]S[table-format=2.2]S[table-format=+1.2]@{}}
\toprule
\textbf{Variant} & {\boldmath$F(g,d)$} & {\textbf{LIBERO} $\uparrow$} & {\textbf{Plus} $\uparrow$} & {$\boldsymbol{\Delta}$\textbf{Plus (pp)} $\uparrow$}\\
\midrule
\multicolumn{5}{@{}l}{\textit{Reference initialization}}\\
No geometry & --- & 92.50 & 38.64 & -0.21\\
Static geometry (baseline) & $0$ & 90.30 & 38.85 & +0.00\\
Bilinear & $g\odot d$ & 92.30 & 39.63 & +0.78\\
\midrule
\multicolumn{5}{@{}l}{\textit{Small-corpus robot-pretrained initialization}}\\
Static geometry (baseline) & $0$ & 97.78 & 67.92 & +0.00\\
Additive & $g+d$ & 97.75 & 67.92 & +0.00\\
Nonlinear additive & $\mathrm{SiLU}(g+d)$ & 97.68 & 67.89 & -0.03\\
Nonlinear geometry & $\mathrm{SiLU}(g)\odot d$ & 97.87 & 67.90 & -0.02\\
Nonlinear dynamics & $g\odot\mathrm{SiLU}(d)$ & 97.58 & 67.72 & -0.20\\
Bilinear & $g\odot d$ & 97.63 & 69.14 & +1.22\\
\bottomrule
\end{tabular*}
\endgroup

\end{table}

Within the robot-pretrained setting, additive coupling matches the static baseline, 67.92\%, and the three nonlinear variants remain within 0.20 points of it, at 67.89\%, 67.90\%, and 67.72\%. The product improves LIBERO-Plus by 1.22 points relative to both the static and additive baselines. Among the tested forms, these mean results favor the product. The zero-factor property in \cref{sec:system} motivates this form.

\begin{figure}[!htbp]
\centering
\paperfigure{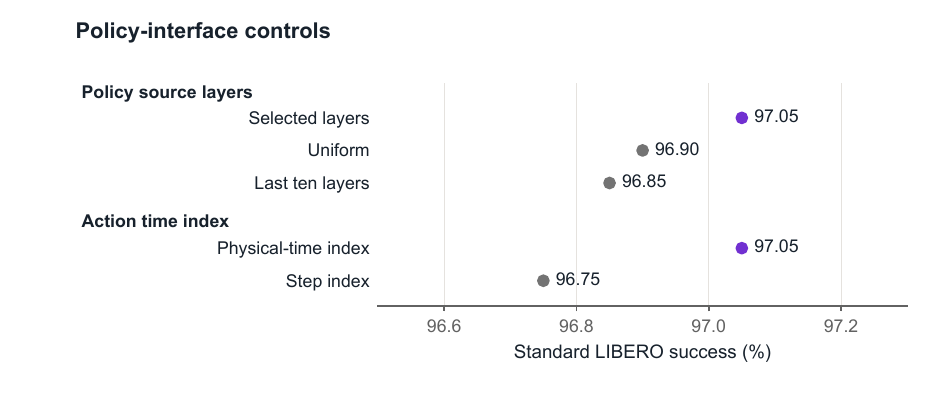}
\caption{\textbf{Policy-interface choices.} Standard LIBERO point estimates for source-layer selection and action indexing, grouped by intervention. Purple identifies the deployed options. The enlarged point-plot range makes the small reported differences visible.}
\label{fig:reader-controls}
\end{figure}

Standard LIBERO varies little across the six robot-pretrained variants, from 97.58\% to 97.87\%, and on this suite multiplicative conditioning is 0.15 points below the static baseline. The geometry variants separate more on the perturbation suite. The two policy-interface choices have small effects on the standard suite (\cref{fig:reader-controls}). The deployed middle-to-late layer selection reaches 97.05\%, compared with 96.90\% for uniformly spaced layers and 96.85\% for the final ten layers, and indexing actions by physical time reaches 97.05\%, compared with 96.75\% for step indexing. The layer-selection differences are 0.15--0.20 points, and the difference between physical-time and step indexing is 0.30 points. With one comparison per choice, these small differences cannot rank policy interfaces in general.

\section{Analysis}
\label{sec:analysis}

This section characterizes the state from which joint training begins, through losses and gradients (\cref{sec:probes}) and attention statistics (\cref{sec:res-attention}), and relates these measurements to the analyses of \cref{app:theory} (\cref{sec:interpretation}). Decoded visual examples are shown in \cref{app:qualitative}.

\subsection{Initialization Losses and Gradients}
\label{sec:probes}
The branch-initialization study compares outcomes after a total budget of 30,000 updates. The probes in this section characterize the state from which S2 begins, and help interpret the differences in downstream outcomes. \cref{tab:probes} reports losses and gradients for all four initialization pairs on eight fixed LIBERO batches, with inputs, masks, positional encodings, and sampled noise held identical across conditions, so that every difference is attributable to the weights. With $R_0$ fixed, replacing $W_0$ by $W_1$ reduces the visual loss from 7.0856 to 0.2089, a 33.9-fold reduction, and the action loss from 1.6769 to 1.6003. Under the unit loss weights of \cref{eq:loss}, the visual term accounts for 98.9\% of the 6.9533 reduction in total loss: nearly all of the change produced by S1 before S2 begins is reflected in the visual objective.

\begin{table}[!htbp]
\centering
\caption{\textbf{Initialization probes on eight fixed LIBERO batches.} $g_w=\nabla_W\mathcal L_w$ and $g_a=\nabla_W\mathcal L_a$ are gradients into the video generation model. The rows distinguish the four weight-source combinations; cross-camera interventions are reported separately in \cref{tab:camera-probes}.}
\label{tab:probes}
\setlength{\tabcolsep}{5pt}\small
\begingroup
\fontsize{9.3}{12}\selectfont
\renewcommand{\arraystretch}{1.08}
\sisetup{group-digits=integer}
\begin{tabular*}{\linewidth}{@{\extracolsep{\fill}}lS[table-format=1.4]S[table-format=1.4]S[table-format=2.3]S[table-format=2.3]S[table-format=-1.3]@{}}
\toprule
\textbf{Initialization} & {\boldmath$\mathcal L_w$} & {\boldmath$\mathcal L_a$} & {\boldmath$\|g_w\|_2$} & {\boldmath$\|g_a\|_2$} & {\boldmath$\cos(g_w,g_a)$}\\
\midrule
$(W_0,R_0)$ & 7.0856 & 1.6769 & 42.734 & 32.366 & 0.103\\
$(W_1,R_0)$ & 0.2089 & 1.6003 & 2.049 & 31.066 & 0.013\\
$(W_0,R_1)$ & 7.0856 & 1.6311 & 42.736 & 23.937 & -0.070\\
$(W_1,R_1)$ & 0.2089 & 1.6000 & 2.049 & 38.166 & -0.044\\
\bottomrule
\end{tabular*}
\endgroup

\end{table}

\begin{figure}[!htbp]
\centering
\paperfigure{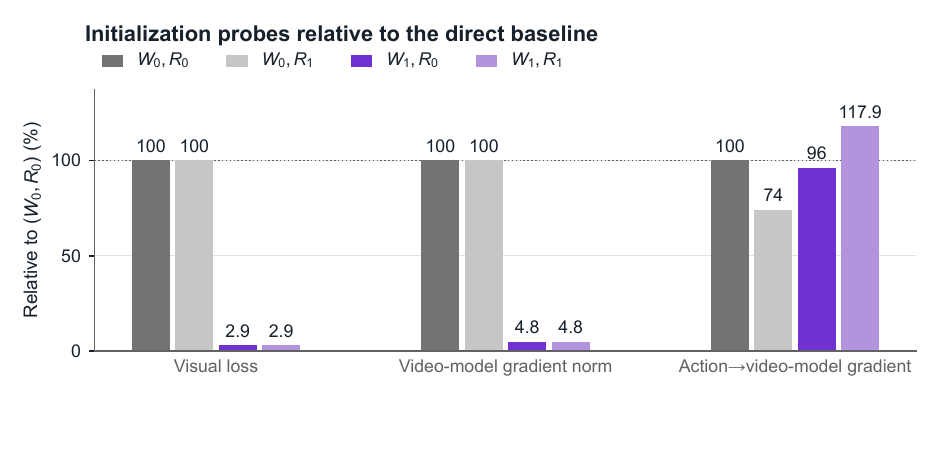}
\caption{\textbf{Loss and gradient probes before S2.} Visual loss, visual-gradient norm, and action-to-video-model gradient norm on eight fixed batches, each normalized to its own $(W_0,R_0)$ value (100\%). Purple shades identify $W_1$ configurations. Labels give the remaining percentage. Absolute losses and gradients, and gradient cosines are retained in \cref{tab:probes}.}
\label{fig:initialization-probes}
\end{figure}

The table is internally consistent with the architecture. The visual loss is identical across policy initializations at a given video-model initialization, 7.0856 at $W_0$ and 0.2089 at $W_1$, and the visual-loss gradient norm is nearly unchanged, because $\mathcal L_w$ does not depend on policy parameters. Policy substitution therefore changes the total loss only through the action term. At $W_0$, replacing $R_0$ by $R_1$ reduces the action loss by 0.0458, from 1.6769 to 1.6311; at $W_1$, by 0.0003. Differences in initial action loss are two orders of magnitude smaller than the change in visual loss, whereas the transfer study shows that downstream differences of tens of points follow video-model initialization. The policy-only configuration begins with the lower action loss of the two $W_0$ pairs and attains the lower final success, a relation between initial loss and final outcome that \cref{eq:spectral} of \cref{sec:theory-transfer} relates to the curvature spectrum of the policy objective.

The gradient statistics describe the optimization problem that S2 inherits. The visual-loss gradient norm falls from 42.734 to 2.049, a reduction of 95.2\%. With $R_0$ fixed, the action-loss gradient into the video generation model changes only from 32.366 to 31.066, so that 96.0\% of it is retained. The ratio $\|g_w\|/\|g_a\|$ accordingly falls twentyfold, from 1.32 to 0.066, and the cosine between the two gradients falls from 0.103 to 0.013. At direct-training initialization, the visual-loss gradient has the larger raw norm; at the start of S2 the visual residual and its gradient have been largely reduced, while the gradient that the action objective sends into the video generation model is nearly unchanged. These probes fit the hypothesis that this initialization makes action updates more effective, although they describe only the starting point of S2 (\cref{fig:initialization-probes}).

Unlike the visual-loss gradient, the action-loss gradient into the video generation model also depends on policy initialization, and the sign of the change differs between the two video-model initializations. At $W_0$, replacing $R_0$ by $R_1$ reduces this norm by 26.0\%, to 23.937; at $W_1$, the same substitution increases it by 22.9\%, to 38.166. In these probes, the inherited policy weights change how much action signal reaches the video generation model, and whether copied S1 layers increase or decrease that signal depends on whether the video generation model has itself been adapted.

The displacement that produces these changes is small. The relative difference $\|W_1-W_0\|_2/\|W_0\|_2$ is 0.01675 and the parameter cosine is 0.99986: the adapted weights are a 1.675\% perturbation almost collinear with the original. A movement of this size reduces one term of the initial loss by more than an order of magnitude while leaving the other nearly unchanged; this change is already present before the first S2 update.

\begin{figure}[!htbp]
\centering
\paperfigure{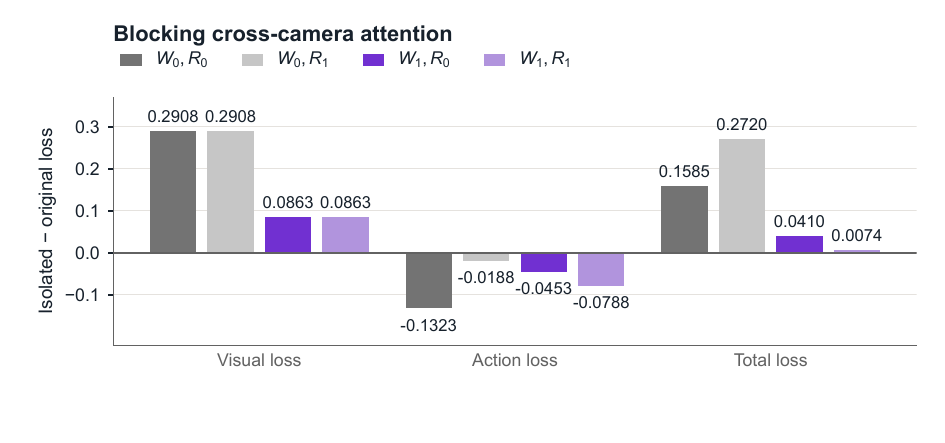}
\caption{\textbf{Effects of blocking cross-camera attention.} Changes in visual, action, and total loss, $\Delta L=L_{\mathrm{isolated}}-L_{\mathrm{original}}$, for four initialization pairs. Positive values indicate increased loss. All values are retained in \cref{tab:camera-probes}. Relative visual-loss changes remain in the text.}
\label{fig:camera-probes}
\end{figure}

\paragraph{Cross-Camera Interaction}
\label{sec:camera}
The frame-causal mask permits tokens from different cameras to attend to one another within a frame, so the representation of each view depends on the others. To test whether cross-view exchange helps both visual prediction and action readout, we block attention between the external and wrist cameras while preserving inputs, latent layout, camera embeddings, positional encodings, validity masks, and noise. Each initialization uses the same eight batches as the initial-state probes. \cref{tab:camera-probes} reports $\Delta_{\mathrm{cam}}\mathcal L_q=\mathcal L_q^{\mathrm{isolated}}-\mathcal L_q^{\mathrm{full}}$ for visual, action, and total loss. Blocking cross-camera attention increases the visual loss in every configuration, by 0.2908 at $W_0$ and by 0.0863 at $W_1$; adaptation thus reduces the absolute increase by 70.3\%. Relative to each configuration's own visual loss, the increases are 4.1\% at $W_0$ and 41.3\% at $W_1$, because adaptation reduces the visual loss itself 33.9-fold. Cross-camera masking therefore produces a larger relative increase over the adapted model's much smaller baseline visual loss (\cref{fig:camera-probes}).

\begin{table}[!htbp]
\centering
\caption{\textbf{Loss changes after removing cross-camera attention.} Measurements on eight fixed LIBERO batches for each initialization. Each entry is isolated loss minus full-attention loss; values are reported to four decimal places.}
\label{tab:camera-probes}
\begingroup
\fontsize{9.3}{12}\selectfont
\sisetup{group-digits=integer}
\begin{tabular*}{\linewidth}{@{\extracolsep{\fill}}lS[table-format=1.4]S[table-format=-1.4]S[table-format=1.4]@{}}
\toprule
\textbf{Initialization} & {\boldmath$\Delta_{\mathrm{cam}}\mathcal L_w$} & {\boldmath$\Delta_{\mathrm{cam}}\mathcal L_a$} & {\boldmath$\Delta_{\mathrm{cam}}\mathcal L$}\\
\midrule
$(W_0,R_0)$ & 0.2908 & -0.1323 & 0.1585\\
$(W_1,R_0)$ & 0.0863 & -0.0453 & 0.0410\\
$(W_0,R_1)$ & 0.2908 & -0.0188 & 0.2720\\
$(W_1,R_1)$ & 0.0863 & -0.0788 & 0.0074\\
\bottomrule
\end{tabular*}
\endgroup

\end{table}

The action loss changes in the opposite direction. Blocking cross-camera attention reduces it by 0.1323, 0.0453, 0.0188, and 0.0788 for $(W_0,R_0)$, $(W_1,R_0)$, $(W_0,R_1)$, and $(W_1,R_1)$, and, unlike the visual-loss response, these values depend on both initializations. The net effect on total loss consequently ranges from 0.2720 at $(W_0,R_1)$ to 0.0074 at $(W_1,R_1)$, a factor of 37 across the four initialization configurations. Cross-view masking therefore affects the two offline losses differently, and the responses depend on the inherited weights. These are offline loss changes.

\subsection{Attention Statistics}
\label{sec:res-attention}
Losses and gradients establish that S1 changes the state from which S2 begins. Attention statistics indicate how the changed representation is used. \cref{tab:attention} compares broad-noise and fixed-noise S1 at their 30,000-update checkpoints, using fixed input samples, fixed future noise, and the same action-denoising step indices (0, 5, and 9), to align the diagnostic inputs. The two runs may differ in more than the S1 noise schedule. For eligible keys $K$ with probabilities $p_j$, group mass is $m_G=\sum_{j\in G}p_j$ and normalized entropy is $H(p)=-\sum_{j\in K}p_j\log p_j/\log|K|$. Action-token top-$k$ mass and entropy use $\widetilde p_j=p_j/m_A$ within the action-token set $A$; other group masses use the shared softmax. Measurements aggregate over tapped layers and action-flow steps 0, 5, and 9.

\begin{table}[!htbp]
\centering
\caption{\textbf{Attention statistics under broad-noise versus S1 with $\sigma_w=1$, at 30k checkpoints with fixed inputs and matched action-denoising steps.} Masses are fractions of the attention distribution.}
\label{tab:attention}
\setlength{\tabcolsep}{5pt}\small
\sisetup{group-digits=integer}
\begingroup
\fontsize{9.3}{12}\selectfont
\renewcommand{\arraystretch}{1.08}
\begin{tabular*}{\linewidth}{@{\extracolsep{\fill}}lS[table-format=1.4]S[table-format=1.4]S[table-format=+1.4]@{}}
\toprule
\textbf{Attention statistic} & {\textbf{Broad-noise S1}} & {\textbf{Fixed} $\sigma_w{=}1$} & {$\boldsymbol{\Delta}$ \textbf{(broad $-$ fixed)}}\\
\midrule
\multicolumn{4}{l}{\textit{Video generation model, visual attention}}\\
Normalized entropy & 0.738 & 0.826 & -0.088\\
Past-frame mass & 0.565 & 0.477 & +0.088\\
Same-frame mass & 0.435 & 0.523 & -0.088\\
Cross-camera mass & 0.227 & 0.234 & -0.007\\
\midrule
\multicolumn{4}{l}{\textit{Policy, action attention}}\\
Top-1 conditional mass & 0.175 & 0.162 & +0.013\\
Top-4 conditional mass & 0.550 & 0.516 & +0.034\\
Normalized entropy & 0.872 & 0.893 & -0.020\\
Total action-token mass & 0.365 & 0.450 & -0.085\\
Dynamic-video mass & 0.612 & 0.501 & +0.110\\
Geometry mass & 0.0235 & 0.0488 & -0.0253\\
\bottomrule
\end{tabular*}
\endgroup

\end{table}

\begin{table}[!htbp]
\centering
\caption{\textbf{Attention statistics across initialization conditions.} All nine statistics for six configurations at 30k total updates. Fixed noise means S1 with $\sigma_w=1$; isolation means full action-gradient isolation. Entropies are normalized; masses are attention fractions. Top-1 and top-4 are conditional shares within action-token attention.}
\label{tab:attention-configs}
\begingroup
\fontsize{9.3}{12}\selectfont
\renewcommand{\arraystretch}{1.08}
\sisetup{group-digits=integer}
\begin{tabular*}{\linewidth}{@{\extracolsep{\fill}}lS[table-format=1.4]S[table-format=1.4]S[table-format=1.4]S[table-format=1.4]S[table-format=1.4]S[table-format=1.4]@{}}
\toprule
\textbf{Attention statistic} & {\boldmath$W_1,R_1$} & {\boldmath$W_1,R_0$} & {\boldmath$W_0,R_1$} & {\boldmath$W_0,R_0$} & {\textbf{Fixed noise}} & {\textbf{Isolation}}\\
\midrule
\multicolumn{7}{@{}l}{\textit{Video model, visual attention}}\\
Normalized entropy & 0.7377 & 0.7370 & 0.7458 & 0.8419 & 0.8257 & 0.8424\\
Past-frame mass & 0.5652 & 0.5659 & 0.4728 & 0.5246 & 0.4772 & 0.5358\\
Cross-camera mass & 0.2267 & 0.2273 & 0.2220 & 0.2594 & 0.2339 & 0.2647\\
\midrule
\multicolumn{7}{@{}l}{\textit{Policy, conditional action attention}}\\
Top-1 conditional mass & 0.1749 & 0.1718 & 0.1617 & 0.1632 & 0.1622 & 0.1459\\
Top-4 conditional mass & 0.5495 & 0.5430 & 0.5171 & 0.5217 & 0.5157 & 0.4820\\
Normalized entropy & 0.8723 & 0.8760 & 0.8937 & 0.8895 & 0.8927 & 0.9117\\
\midrule
\multicolumn{7}{@{}l}{\textit{Policy, total allocation}}\\
Action-token mass & 0.3649 & 0.3677 & 0.4506 & 0.4700 & 0.4497 & 0.3993\\
Dynamic-video mass & 0.6116 & 0.6085 & 0.4928 & 0.4722 & 0.5015 & 0.5447\\
Geometry mass & 0.0235 & 0.0238 & 0.0565 & 0.0578 & 0.0488 & 0.0560\\
\bottomrule
\end{tabular*}
\endgroup

\end{table}

In the video generation model, broad-noise S1 makes visual attention more selective and redirects it along the temporal axis. Normalized entropy falls from 0.826 to 0.738, and past-frame mass rises from 0.477 to 0.565, an exact exchange with same-frame mass because the two partition the distribution. Cross-camera mass changes by only 0.007, from 0.234 to 0.227. Broad-noise adaptation therefore reorganizes attention along the temporal axis, drawing more on preceding frames, while the allocation between cameras changes little.

In the policy, attention concentrates on fewer action tokens. Within the distribution normalized over action tokens, top-1 mass rises from 0.162 to 0.175 and top-4 mass from 0.516 to 0.550, where top-$k$ mass sums the $k$ largest probabilities, and normalized action entropy decreases from 0.893 to 0.872. Total action-token mass across all token groups moves from 0.450 to 0.365. The policy thus attends to action tokens more selectively and reallocates attention toward the video generation model: dynamic-video mass rises from 0.501 to 0.612, while geometry mass moves from 0.0488 to 0.0235.

The shift in geometry attention is associated with the interaction found in \cref{sec:factorial}. Under broad-noise adaptation the policy's geometry mass decreases by 51.8\%, from 0.0488 to 0.0235, while its dynamic-video mass increases from 0.501 to 0.612. This shift is consistent with the hypothesis that adapted video-model features overlap with information supplied by geometry. The factorial interaction point estimate is $-1.61$ points, but that study and the noise comparison vary different factors, so the attention shift accompanies the interaction without being shown to cause it.

\paragraph{Attention Under Different Initializations}
\label{sec:res-attention-transfer}
Extending the same diagnostic to all six configurations provides a complementary diagnostic for the branch-transfer results of \cref{sec:transfer}. The two configurations with an adapted video generation model, $(W_1,R_0)$ and $(W_1,R_1)$, agree to within 0.0007 in visual entropy, 0.0007 in past-frame mass, 0.0031 in dynamic-video mass, and 0.0003 in geometry mass (\cref{tab:attention-configs}). The visual entropy of the adapted video-model pair differs by only 0.0007, while its separation from the direct and gradient-isolation configurations is about 0.104. Thus, for the adapted video generation model, the two policy initializations produce similar values of the reported attention statistics. With the original video generation model, the two policy initializations differ more.

Policy-only transfer shows that this pattern depends on several statistics together. Its visual entropy, 0.7458, is close to the 0.7370--0.7377 of the video-model-transfer pair and well below the 0.8419 of direct training, whereas its past-frame mass of 0.4728 and dynamic-video mass of 0.4928 lie below both video-model-transfer configurations, and its geometry mass of 0.0565 matches the unadapted group. Copied S1 layers in the policy thus reproduce the lower visual entropy without the temporal and dynamic reallocation that accompanies adaptation of the video generation model, and this configuration has the lowest success of the four. The attention pattern associated with the highest success is therefore the joint pattern of lower entropy, greater past-frame mass, greater dynamic-video mass, and smaller geometry mass, which, among the six evaluated configurations, appears only in the two with a broad-noise-adapted video generation model.

\subsection{Interpretation}
\label{sec:interpretation}
S1 and S2 specialize the pretrained prior toward different objectives. S1 learns to predict target-domain visual change across noise levels; S2 learns to extract action-relevant information from the resulting representation while continuing to receive visual supervision. We refer to these as visual specialization and action-oriented specialization. The probes of \cref{sec:probes} are consistent with the distinction: after S1 the visual residual and its gradient are small while the action-derived gradient into the same parameters is nearly unchanged. The attention statistics of \cref{sec:res-attention} provide representational evidence for it, with entropy falling in both branches and attention shifting between token groups. The branch-initialization study shows what this means in practice: the branch that benefits from a history of visual specialization is the video generation model, and the policy attains its highest perturbation-suite score when initialized from the original weights.

The measurements of \cref{sec:probes} establish what holds at the adapted state $C$ of \cref{sec:discussion}: a visual loss reduced 33.9-fold, an action-derived gradient retained at 96.0\%, and a parameter displacement of 1.675\%. \cref{app:theory} offers possible accounts of why $C$ is a better starting point for joint training and why its weights favor the video generation model (\cref{sec:theory-noise,sec:theory-tracking,sec:theory-transfer}), and when the cost of reaching $C$ is recovered (\crefrange{sec:theory-area}{sec:theory-cost}). Two of these analyses relate to the experiments.

The noise-level analysis of \cref{sec:theory-noise} offers a possible interpretation of the fixed-noise control of \cref{sec:res-noise}. Supervising only at $\sigma_w=1$ trains the video generation model on a functional, the conditional mean, that a deterministic future and a multimodal future can share; supervising across $\sigma$ trains it on a family of posterior means that distinguishes them. This motivates the hypothesis that broad-noise supervision may be useful when futures are uncertain. Testing it would require results by perturbation category; the 20.80-point gap between the two noise schedules is an aggregate LIBERO-Plus comparison.

The tracking bound of \cref{sec:theory-tracking} identifies distinct roles for the two initializations. The bound separates the initial tracking error $e_0$ from subsequent video-model parameter drift. Both initializations can affect these quantities, so the best choice for one branch may depend on the other. It also separates three ways in which S1 could accelerate joint training: by improving the starting point, by slowing the motion of the target, or by making the action objective easier to descend through a larger local curvature $\mu$. The initialization probes measure losses and raw gradients; policy-optimum distances, subsequent drift, and local curvature remain unmeasured.

\section{Related Work}
\label{sec:related}

\paragraph{Video--Action Models}
Vision--language--action (VLA) models and video--action models (VAMs) are two prominent foundation-model approaches to generalist robot control. VLA policies inherit semantic representations from vision--language pretraining~\cite{rt2,openvla}, whereas VAMs connect predictive video representations to action learning~\cite{wamsurvey}. Earlier work learned behavior through latent dynamics models~\cite{worldmodels,dreamerv3} or future-measurement prediction~\cite{dfp}, and human video supplied pretrained visual representations~\cite{r3m}. Video generators have also served as visual planners~\cite{unipi,avdc}, interactive simulators~\cite{unisim,genie,cosmoswfm}, and synthetic data engines~\cite{dreamgen}. Building on video-based policies such as GR-1 and GR-2~\cite{gr1,gr2}, recent video-based policies initialize from web-scale video generators~\cite{cosmospolicy,videovla,mimicvideo,dreamzero,lingbotva}, pretrain on robot or egocentric video~\cite{genieenvisioner,gigaworldpolicy,gigaworldpolicy05,internw0}, train generative and control components from scratch on manipulation data~\cite{geact2}, or unify world modeling and control in an omnimodal model~\cite{cosmos3}. Their action interfaces include deriving actions from generated future states~\cite{unipi,mimicvideo,geact2}, planning with predicted states and values~\cite{cosmospolicy}, and jointly predicting video and actions~\cite{uva,uwm,motus,videovla,dreamzero,lingbotva}. Another route reads internal video features without decoding future frames for control. Diffusion models provide features for recognition and correspondence~\cite{ddae,dift,ldae}, while representation alignment can improve generative training~\cite{repa}. In robot policies, VPP extracts predictive features in a single pass~\cite{vpp}; EnerVerse and DiT4DiT use first-step or intermediate denoising features~\cite{enerverse,dit4dit}; Genie Envisioner decodes robot-video latents into actions~\cite{genieenvisioner}; and InternW0 reuses layerwise keys and values~\cite{internw0}. Fast-WAM and GigaWorld-Policy retain future-video supervision during training while decoding only actions at deployment~\cite{fastwam,gigaworldpolicy,gigaworldpolicy05}. Related work improves this interface through reused future representations~\cite{fasterwam}, shallow action modules~\cite{dot}, latent future memory~\cite{forewam}, asynchronous context reuse~\cite{aha}, and distilled geometric priors~\cite{meco}. ZERITH V2.5 uses an action policy over intermediate video-model keys and values. Its contribution concerns the state of this representation when action learning begins, complementing work on the readout interface.

\paragraph{Policy Training and Adaptation}
Coordinating a pretrained backbone with a newly attached action module is a central training choice. In VLA systems, $\pi_{0.5}$ introduces a flow-matching expert after discrete-action pretraining~\cite{pi05}; knowledge insulation limits the expert's gradients into the backbone~\cite{ki}; and Wall-OSS-0.5 and GigaBrain-0.7 jointly optimize vision--language and action objectives in one stage~\cite{walloss05,gigabrain07}. Video-based policies likewise adapt predictive models before or alongside action learning~\cite{gr2,vpp,genieenvisioner,gigaworldpolicy,vpp2}. DreamGen specializes video models before generating policy-training data~\cite{dreamgen}; UVA reports video pretraining followed by video--action co-training~\cite{uva}; and Motus and GE-Act 2.0 use multi-stage pipelines~\cite{motus,geact2}. Video Policy compares staged and joint schedules, but gives the staged configuration additional updates~\cite{videopolicy}. Controlled studies further distinguish benefits on familiar tasks from robustness under distribution shift~\cite{wmimprove}, associate pretrained representation geometry with downstream robustness~\cite{reppriors}, and find that robot specialization improves prediction and control in their evaluated settings~\cite{greenwam}. These questions connect to intermediate domain adaptation~\cite{ulmfit,dontstop}, staged video-model training~\cite{svd}, and the ordering of training signals~\cite{curriculum}. Transfer studies also examine task divergence~\cite{yosinski}, feature distortion when a new head and backbone are fine-tuned together~\cite{lpft}, and gradient scaling between pretrained and new components~\cite{defrcn}. For diffusion and flow models, adaptation additionally depends on the allocation of supervision across noise levels: EDM and Stable Diffusion~3 study noise distributions and weighting~\cite{edm,sd3}, while Min-SNR and P2 reweight denoising losses for optimization or perceptual learning~\cite{minsnr,p2}. This differs from choosing the noise level at which a policy reads features, as in DiT4DiT and UWM~\cite{dit4dit,uwm}. ZERITH V2.5 studies these choices at matched demonstrations and total updates. It adapts the video generation model across the full noise range before joint training, although the policy reads it only at $\sigma_w=1$. In our controls, endpoint-only adaptation does not reproduce the gain, and the gain follows the adapted video-model weights. The analysis in \cref{sec:theory-noise} relates intermediate-noise supervision to posterior structure beyond the endpoint's conditional mean. Our schedule also restores action gradients after adaptation: withholding them throughout joint training performs worse than direct training in the evaluated setting.

\paragraph{Action Prediction and Conditioning}
Robot policies differ in both how they generate actions and what information conditions them. Vision--language--action systems transfer pretrained multimodal knowledge to control~\cite{rt2,openvla,pi0,pi05,pi07,gr00t,gr3,geminirobotics,geminirobotics15,walloss05,qwenmanip,gigabrain07}, alongside generalist controllers trained on robot trajectories without vision--language initialization~\cite{rt1,octo,rdt}. Large multi-embodiment datasets support these policies~\cite{oxe,agibot}. RT-2 and OpenVLA decode discrete action tokens~\cite{rt2,openvla}, and FAST compresses action sequences before tokenization~\cite{fast}. CogACT uses a diffusion action module~\cite{cogact}, following diffusion policies for visuomotor control~\cite{diffusion_policy}; $\pi_0$ and $\pi_{0.5}$ use flow-matching experts~\cite{pi0,pi05}; and GR00T~N1 combines a vision--language model with a diffusion transformer~\cite{gr00t}. OpenVLA-OFT studies parallel decoding, action chunking, and other fine-tuning choices~\cite{openvlaoft}. Predictive conditioning includes latent actions learned from video~\cite{lapa,univla}, future observations or subgoal images~\cite{seer,cotvla,pi07}, aligned future embeddings~\cite{flare}, and foresight tokens supervised by a frozen video generator~\cite{internvlaa15}. ZERITH V2.5 uses flow-based action prediction~\cite{flowmatching,pi0} and shared-softmax attention over modality-specific parameter sets, following Mixture-of-Transformers~\cite{mot}. Video-model, geometry, and action memory thus enter one attention operation without merging their weights. Geometry can condition policies through explicit 3D observations~\cite{dp3}, pretrained geometric features~\cite{vodp}, or a 3D foundation-model backbone~\cite{e05}. Our geometry branch instead modulates exported video-model values with a multiplicative residual, related in role to feature-wise modulation~\cite{film} and built from low-rank, zero-initialized components~\cite{lora,controlnet}. Holding this conditioning architecture fixed within training-schedule comparisons separates the effect of the recipe from that of the conditioning operator on LIBERO~\cite{libero} and LIBERO-Plus~\cite{plus}.

\section{Conclusion, Limitations, and Future Directions}
\label{sec:conclusion}

\subsection{Conclusion}
This report asked what training a video generation model needs before an action policy can use its predictive priors effectively, with the demonstrations and the total number of updates held fixed. Across controlled comparisons, two conditions matter: how predictive training is supervised, and which module receives the adapted weights. ZERITH V2.5 summarizes the resulting recipe as \emph{adapt the visual world, then scale the action policy}. Under a matched 30,000-update budget on the same demonstrations, devoting the first 3,000 updates to video-model-only adaptation improves LIBERO-Plus success by 17.55--19.16 percentage points across three geometric-conditioning variants with three seeds each, compared with 0.68--2.29 points from changing the operator within the same factorial study. The advantage also extends beyond a single endpoint. At matched Stage~2 update counts after the first few thousand updates, the policy built on the adapted video generation model leads direct training by 8 to 10 points at every recorded checkpoint; on the total-update axis, the staged schedule at 30,000 updates remains 18.18 points ahead of direct training at 50,000 updates.

Ablations and initialization probes show where this gain comes from and what it depends on. Crossing the two initialization sources shows that the benefit moves specifically with the adapted video-model weights: transferring them to the video generation model alone adds 19.67 points on LIBERO-Plus, whereas copying adapted layers into the action policy alone lowers success by 3.94 points. Two further controls narrow these findings: restricting Stage~1 adaptation to the pure-noise endpoint read by the policy, or withholding action gradients from the video generation model throughout direct training, each reduces success below standard direct training. These results support the roles of broad-noise adaptation and action-gradient feedback in the evaluated configurations. At the start of Stage~2, a 1.675\% relative displacement of the video-model weights has reduced the visual-loss gradient norm by 95.2\% while retaining 96.0\% of the action-loss gradient into the video generation model, a starting state we hypothesize makes subsequent action updates more effective. With pretraining on 10,000 hours of cleaned and selected robot data, ZERITH V2.5 reaches 98.95\% on standard LIBERO and 93.70\% on LIBERO-Plus, and on a real robot it matches or exceeds the compared baselines on pick and place and bin lining. For the setting we evaluated, the findings suggest three practical guidelines for video--action policies: adapt the video generation model to the target visual domain across the full noise range before introducing the action policy; initialize the video generation model and the action policy independently, assigning the adapted weights to the video generation model; and allow action gradients to refine the video-model representation during joint training.

\subsection{Limitations}
\label{sec:limitations}
\paragraph{Scope.} All controlled comparisons use one video generation model (Wan2.2-TI2V-5B), one adaptation length ($\tau=3{,}000$ updates), and the LIBERO benchmark family. Whether the gain carries over to other backbones, task domains, or adaptation lengths is untested, so the guidelines above apply to this setting. The real-robot evaluation compares complete systems on two tasks, and real-robot trials involve some randomness. Baseline scores in \cref{tab:comparison} are reported numbers, each from its own evaluation protocol.

\paragraph{Budget crossover.} On LIBERO-Plus, staged training trails direct training at 5,000 total updates and leads at every evaluated budget from 30,000 updates onward. The crossing lies between these budgets, but the evaluated checkpoints cannot pin it down. The staged 8,000-update checkpoint exceeds the direct 10,000-update checkpoint by only 0.14 points, and direct training was not evaluated at 8,000 updates. Placing the crossover in the 5,000--8,000-update interval therefore assumes that the direct score at 8,000 updates does not exceed its 10,000-update score. The broader 5,000--13,000-update interval ($1.7\tau$--$4.3\tau$) also rests on unmeasured equal-budget points. Update savings relative to 50,000 direct updates refer only to the evaluated budgets.

\paragraph{Mechanism.} The initialization probes describe only the state at the start of Stage~2. The claim that this state makes action updates more effective remains a hypothesis; testing it would require an intervention. The analyses in \cref{app:theory} are local or illustrative models, and no parameter of the optimal-length model is fitted to data. \cref{app:limitations} lists seed counts, statistical precision, and the remaining limitations by topic.

\subsection{Future Directions}
We see two directions for future work. The first concerns generality: evaluating staged visual grounding across additional video backbones and task families, and measuring how the budget crossover and downstream gain vary with the adaptation length $\tau$, would test how broadly the recipe applies and turn the trade-off analysis of \cref{sec:theory-tau} into a practical rule for selecting $\tau$. The second concerns multi-task reuse: the amortization threshold of \cref{eq:reuse} indicates that the cost of Stage~1 adaptation can be shared across many downstream policies, making a domain-adapted video generation model a natural candidate for a shared, task-independent foundation in multi-task robot learning.

\FloatBarrier

\clearpage
\phantomsection
\section*{\hypertarget{report-zlab}{Contributors and Acknowledgments}}
\addcontentsline{toc}{section}{Contributors and Acknowledgments}
\label{sec:contributors}
\ReportCreditHeading{Contributors}

\ReportCreditGroup{Model development and pretraining}{Ruby, Epoch, Guoping Pan}

\ReportCreditGroup{Data}{Orion, Luke, Guoping Pan, Ruby, Chenxi Han}

\ReportCreditGroup{Post-training and evaluation}{Guoping Pan, Ruby, Orion, Bin, Dusk, Lyra, Cal, Hubert, Alex}

\ReportCreditGroup{Real-robot deployment}{Chenxi Han, Alvin, Jiahe Han, Tao Zhang}

\ReportCreditGroup{Manuscript, analysis, and visualization}{Epoch, Chenxi Han, Yi Cheng, Guoping Pan, Dusk, Hang Liu, Shoujie Li, Yizhe Li, Ruby}

\ReportCreditHeading{Acknowledgments}
\noindent We thank Sirius for project coordination and Storm for computing and data-processing infrastructure support. We also thank the data-collection and on-site teams for preparing robot demonstrations and supporting experiments, and the developers and maintainers of the open-source models, datasets, and benchmarks used in this work.

\clearpage

\FloatBarrier\Needspace{10\baselineskip}
\bibliographystyle{zerith-report}
\bibliography{references}

\FloatBarrier\Needspace{12\baselineskip}
\appendix
\crefalias{section}{appendix}
\crefalias{subsection}{subappendix}
\addtocontents{toc}{\protect\setcounter{tocdepth}{1}}

\ReportAppendixSection{Theoretical Analysis}{app:theory}

This appendix collects the analyses summarized in \cref{sec:discussion,sec:interpretation}. Their assumptions and scope are discussed in \cref{app:limitations}.

\subsection{Noise-Level Coverage}
\label{sec:theory-noise}
We first compare the population objectives at different noise levels. This analysis concerns squared-error flow matching with Gaussian noise independent of the clean target conditional on the observation context $c$.

Let $Y=\epsilon-x^\star$ and $x_\sigma=(1-\sigma)x^\star+\sigma\epsilon$. The squared-error minimizer is the conditional expectation of $Y$. For $\sigma>0$, substituting $\epsilon=[x_\sigma-(1-\sigma)x^\star]/\sigma$ gives
\begin{equation}
v^*(x,c,\sigma)=\frac{x-\mathbb E[x^\star\mid x_\sigma=x,c,\sigma]}{\sigma}.
\label{eq:optvel}
\end{equation}
At $\sigma=1$, the future input $x_1=\epsilon$ is independent of the target conditional on $c$, and \cref{eq:optvel} becomes
\begin{equation}
v^*(\epsilon,c,1)=\epsilon-\mathbb E[x^\star\mid c].
\label{eq:endpoint}
\end{equation}
The endpoint optimum depends on the target distribution through its conditional mean. For example, let $P$ assign equal probability to $x^\star=\pm1$, and let $Q$ assign probability one to $x^\star=0$. Both distributions have mean zero and therefore the same endpoint velocity $v^*(\epsilon,1)=\epsilon$. At intermediate noise, however, $\mathbb E_P[x^\star\mid x_\sigma=x]=\tanh((1-\sigma)x/\sigma^2)$, while $\mathbb E_Q[x^\star\mid x_\sigma=x]=0$. Their optimal velocity fields consequently differ for $0<\sigma<1$.

A local Jacobian analysis describes the associated parameter constraints. At fixed parameters, let $J_\sigma$ be the Jacobian of the predicted velocity with respect to those parameters, and define $G_p=\mathbb E[J_\sigma^\top J_\sigma]$, with the expectation taken over training inputs, Gaussian noise, and $\sigma\sim p$. For any parameter direction $d$, $d^\top G_pd=\mathbb E\|J_\sigma d\|^2$,
\begin{equation}
\ker G_p=\{d:J_\sigma d=0\ \text{almost surely under the sampling distribution}\},
\label{eq:kernel}
\end{equation}
because a nonnegative random variable has zero expectation only when it is zero almost surely. A parameter direction is constrained whenever $J_\sigma d$ is nonzero on a set of positive sampling probability. A direction therefore remains unconstrained only if every sampled noise level fails to constrain it, and noise levels that constrain complementary directions shrink the unidentified set jointly. Broader noise sampling adds constraints exactly when it exposes directions that are unobserved at a fixed noise level. S1 uses $p_1(\sigma)=0.1\delta_0+0.9\,\mathrm{Unif}(0,1)$, where $\delta_0$ is a point mass at zero, so the clean atom and the interval interior can impose complementary constraints beyond endpoint supervision.

\subsection{Policy Tracking Under Video-Model Parameter Updates}
\label{sec:theory-tracking}
Let $\phi$ and $\theta$ denote the video-model and policy parameters, respectively, and let $f(\theta,\phi)$ be the expected action loss with other trainable components held fixed for this local analysis. We analyze how changes in $\phi$ affect tracking of a local policy optimum $\theta^*(\phi)$. Assume that $f$ is twice continuously differentiable in a convex neighborhood containing the iterates and the relevant optima, with $0<\mu\leq L$, $\mu I\preceq H_{\theta\theta}\preceq LI$ and $\|H_{\theta\phi}\|\leq M$. Here $H_{\theta\theta}$ and $H_{\theta\phi}$ are the policy Hessian and mixed Hessian, and norms are Euclidean or induced operator norms. The optimum map is locally Lipschitz with constant $K=M/\mu$. For deterministic policy updates $\theta_{k+1}=\theta_k-\eta_r\nabla_\theta f(\theta_k,\phi_k)$ with $0<\eta_r\leq1/L$, define $e_k=\theta_k-\theta^*(\phi_k)$. Gradient-descent contraction and the Lipschitz bound give
\begin{equation}
\|e_{k+1}\|\leq(1-\eta_r\mu)\|e_k\|+K\|\phi_{k+1}-\phi_k\|.
\label{eq:tracking}
\end{equation}
Writing $q=1-\eta_r\mu$ and $\Delta\phi_j=\phi_{j+1}-\phi_j$, iteration yields
\begin{equation}
\|e_k\|\leq q^k\|e_0\|+K\sum_{j=0}^{k-1}q^{k-1-j}\|\Delta\phi_j\|.
\label{eq:unroll}
\end{equation}
The first term measures the initial distance from the policy optimum; the second accumulates changes in video-model parameters. If $\|\Delta\phi_k\|\leq D_\phi$ at every update, then $\limsup_k\|e_k\|\leq KD_\phi/(\eta_r\mu)$. Under these assumptions, smaller video-model parameter updates reduce the upper bound on policy-tracking error.

The initialization probes measure losses and raw gradients, not policy-optimum distances or subsequent video-model parameter drift. The smaller visual gradient motivates a possible tracking-based explanation, but its effect on AdamW updates is not measured here. AdamW steps depend on gradient history, preconditioning, and weight decay, so the 95.2\% reduction in the initial raw visual-loss gradient cannot be read as the same reduction in the drift term of \cref{eq:unroll}.

\subsection{Why Adapted Weights Favor the Video Generation Model}
\label{sec:theory-transfer}
S1 minimizes the visual objective, and nothing in that objective constrains how the same weights behave when they occupy the policy. Copying video-model layers into the policy changes their input distribution, the semantics of the tokens they process, their attention connectivity, their supervision, and the modules adjacent to them. Whether the displacement $\delta=R_1-R_0$ helps the policy is therefore a property of the action objective at the policy, and a local analysis makes this precise.

Fix the video-model initialization $W_i$ and write the action objective at the policy as $\mathcal A_i(\theta)$, with gradient $g_i$ and Hessian $H_i\succ0$ at $\theta_0=R_0$. In a neighborhood on which $\mathcal A_i$ is quadratic with minimizer $\theta_i^*$, the transfer helps at initialization exactly when the displaced policy lies closer to the optimum in the metric of the curvature,
\begin{equation}
\mathcal A_i(\theta_0+\delta)<\mathcal A_i(\theta_0)
\iff
\|\theta_0+\delta-\theta_i^*\|_{H_i}<\|\theta_0-\theta_i^*\|_{H_i}.
\label{eq:transfer-sign}
\end{equation}
S1 shapes the visual objective, while $g_i$ and $H_i$ are set by the action objective at the policy, so a single $\delta$ can improve the video generation model and move the policy away from its optimum at the same time. \cref{eq:transfer-sign} gives a local loss-based account of the downstream branch asymmetry in \cref{sec:transfer}.

Within this quadratic gradient-flow model, finite-time loss depends on the spectrum of $H_i$ and the error components, beyond the initial loss. Under gradient flow the policy error decays independently along each eigendirection, so with eigenvalues $\lambda_j$ and initial error components $e_j$,
\begin{equation}
\mathcal A_i(t)-\mathcal A_i^*=\tfrac12\textstyle\sum_j\lambda_j\,e_j^2\,e^{-2\lambda_jt}.
\label{eq:spectral}
\end{equation}
Error along high-curvature directions is removed quickly, whereas error along low-curvature directions persists. Different distributions of error across these directions can lead to different finite-time behavior despite similar initial losses. \cref{eq:spectral} therefore gives a possible local explanation of loss trajectories; the residual spectrum is not measured, and the equation does not establish an ordering of final closed-loop success.

The video-model initialization also determines what a good policy displacement is. Writing the policy's best response as $\theta^*(\phi)=\arg\min_\theta\mathcal A(\phi,\theta)$, the implicit-function theorem gives
\begin{equation}
\frac{d\theta^*}{d\phi}=-H_{rr}^{-1}H_{rw},
\label{eq:best-response}
\end{equation}
where $H_{rw}$ is the mixed policy--video-model curvature. Changing the video generation model can shift the policy optimum; the benefit of a copied displacement depends on its effect on the policy's current error under the local curvature metric. The value of $R_1$ can therefore depend on which video generation model it is paired with. A related observation in \cref{sec:probes} is that, paired with the adapted video generation model, the copied policy raises the action-to-video-model gradient by 22.9\%; this does not measure the optimum shift.

These results also turn the binary choice between $R_0$ and $R_1$ into a continuous one. For $\delta\ne0$, along $R_\lambda=R_0+\lambda\delta$ the quadratic model has the unique unconstrained optimum
\begin{equation}
\lambda^*=-\frac{g_i^\top\delta}{\delta^\top H_i\,\delta},
\label{eq:interp}
\end{equation}
If $\lambda$ is restricted to $[0,1]$, this solution is clipped to that interval. Layer-wise interpolation is a possible extension, but cross-layer curvature prevents treating each layer as an independent optimum in general.

\subsection{Crossover for General Learning Curves}
\label{sec:theory-area}
Let $T$ denote training budget measured in optimizer updates, and let $q_D(T)$ and $q_S(T)$ denote direct and staged performance under the same evaluation protocol. Suppose the curves are absolutely continuous after the adaptation budget $\tau$, with derivatives $r_D$ and $r_S$ and a performance deficit at the end of S1, $D_\tau=q_D(\tau)-q_S(\tau)>0$. Then
\begin{equation}
\Delta(T):=q_S(T)-q_D(T)=-D_\tau+\int_\tau^T\!\big[r_S(u)-r_D(u)\big]\,du.
\label{eq:area}
\end{equation}
A crossing occurs when the integral equals $D_\tau$. If $g=r_S-r_D>0$ almost everywhere, the cumulative integral is strictly increasing. A unique finite crossing exists exactly when
\begin{equation}
\int_\tau^\infty g(u)\,du>D_\tau,
\qquad\text{and then}\qquad
\int_\tau^{T^*}g(u)\,du=D_\tau.
\label{eq:existence}
\end{equation}
This follows from continuity and strict monotonicity of the cumulative integral. The crossover is thus the budget at which the accumulated additional progress of the staged route equals the deficit incurred during S1, and the statement assumes no functional form for either curve. If the cumulative advantage converges to a value below $D_\tau$, the staged schedule does not overtake direct training at any budget.

\cref{eq:area} provides an interpretation of continuous learning curves, while the experiments record only selected checkpoints. The staged deficit is 17.06 points at 5,000 total updates. Its 8,000-update score exceeds the direct 10,000-update score, but no direct 8,000-update measurement locates an equal-budget crossing. At later endpoints the staged advantage decreases from 19.16 to 16.68 points between 30,000 and 50,000 updates. This finite difference does not determine the sign of $g$ between evaluations or show that the endpoints belong to one continuous run. If the rate ordering changes more than once in the model, the sign of the cumulative difference in \cref{eq:area} determines which schedule is ahead at each budget.

Two further properties follow directly. A differentiable, strictly increasing transformation $h$ of the performance metric preserves the crossing location, since $q_S=q_D$ if and only if $h(q_S)=h(q_D)$, but multiplies both rates by $h'$; success rate and its logit, where defined, can therefore display different slopes while agreeing on $T^*$. The crossing is also defined relative to a budget axis: update count, token exposures, and execution time yield different values of $\tau$ and hence different crossovers, which is why \cref{sec:res-crossover} reports one axis at a time.

\subsection{Sensitivity of the Linear Crossover}
The dependence of the crossover on each quantity follows by differentiating \cref{eq:crossover}:
\begin{equation}
\begin{aligned}
\frac{\partial T^*}{\partial r_{AC}}&=-\frac{\tau}{r_{CB}-r_{AB}},\quad
\frac{\partial T^*}{\partial r_{AB}}=\frac{\tau(r_{CB}-r_{AC})}{(r_{CB}-r_{AB})^2},\quad
\\
\frac{\partial T^*}{\partial r_{CB}}&=-\frac{\tau(r_{AB}-r_{AC})}{(r_{CB}-r_{AB})^2},\quad
\frac{\partial T^*}{\partial\tau}=\frac{r_{CB}-r_{AC}}{r_{CB}-r_{AB}}.
\end{aligned}
\label{eq:sensitivity}
\end{equation}
The crossover therefore moves earlier when S1 itself advances the control metric, when the post-adaptation rate is higher, or when the same intermediate state is reached with a shorter adaptation stage. Because the second and third derivatives scale with $(r_{CB}-r_{AB})^{-2}$, the crossover is most sensitive to the rates precisely when the post-adaptation advantage is small.

For this illustrative model, assume $r_{AC}=0$ for a hypothetical common control probe during S1; this rate is not measured. Writing $m=T^*/\tau$, \cref{eq:crossover} then inverts to
\begin{equation}
\frac{r_{CB}}{r_{AB}}=\frac{m}{m-1}.
\label{eq:inversion}
\end{equation}
Conditioning on the exploratory interval $5{,}000<T^*<8{,}000$ with $\tau=3{,}000$ gives $1.67<m<2.67$ and hence $1.6<r_{CB}/r_{AB}<2.5$. This model-based rate ratio inherits the unmeasured-point assumptions described in \cref{sec:res-crossover}.

\subsection{Budget Required for a Target Performance}
For target performance $q^\dagger$ above the crossing and within the linear regime, let $Q=q^\dagger-q_0$. \cref{eq:curves} gives
\begin{align}
T_D(q^\dagger)&=Q/r_{AB},\nonumber\\
T_S(q^\dagger)&=\tau+(Q-r_{AC}\tau)/r_{CB}.
\label{eq:target-budget}
\end{align}
Their difference is
\begin{equation}
T_D-T_S=\frac{r_{CB}-r_{AB}}{r_{AB}r_{CB}}\,Q-\frac{r_{CB}-r_{AC}}{r_{CB}}\,\tau.
\label{eq:saving}
\end{equation}
The saving grows linearly in the target and decreases linearly in the adaptation cost. Substituting \cref{eq:crossover} shows that the two efficiency criteria coincide: $T_D-T_S>0$ holds exactly when $T_D(q^\dagger)>T^*$, so the staged schedule reaches a target with a smaller budget precisely when the budget that direct training requires for that target exceeds the crossover. As the target grows, the relative saving approaches a limit determined by the rates alone,
\begin{equation}
\lim_{Q\to\infty}\frac{T_D-T_S}{T_D}=1-\frac{r_{AB}}{r_{CB}},
\label{eq:limit-saving}
\end{equation}
because the fixed cost $\tau$ is amortized over an increasing budget. Conditional on the exploratory interval used in \cref{eq:inversion}, the model limit lies between 37.5\% and 60\%. The checkpoint comparisons of \cref{tab:matched} give update reductions of 56.7\% against the 30,000-update direct endpoint and 74.0\% against the 50,000-update checkpoint. The constant-rate model does not describe the full observed learning curves: direct scores vary little across 20,000--50,000 updates, while staged scores are non-monotone. Its limiting saving is therefore a model consequence.

Under the same linear-model assumptions, the expressions admit an idealized data-size interpretation. If $N$ fixed-length training examples are processed for $e$ complete passes at global batch size $b$, then $T=eN/b$ optimizer updates, ignoring batch rounding. With $e$, $b$, the adaptation budget, and the rates fixed, the corresponding threshold is $N^*=bT^*/e$: direct training is preferable below it and staged training above it within the model. Here $N$ counts training examples or windows, not variable-length demonstrations; this sample threshold is not measured.

\subsection{Optimal Adaptation Length}
\label{sec:theory-tau}
The preceding results take $\tau$ as given. Treating it as a decision variable in an illustrative error model describes when staging should be used and how long S1 should last under that model. Fix the total budget $T$ and let $\tau\in[0,T]$ be the adaptation length. Model the action error after training as
\begin{equation}
E(T,\tau)=E_C\exp\{-\lambda(\tau)(T-\tau)\},
\label{eq:error-model}
\end{equation}
where $E_C$ is held fixed as the starting S2 error in this illustrative model and $\lambda(\tau)$ is the post-adaptation rate of improvement. Suppose that adaptation increases this rate with diminishing returns, $\lambda(\tau)=\lambda_0+\Delta\lambda\,(1-e^{-a\tau})$ with $\lambda_0,\Delta\lambda,a>0$. Minimizing $E$ is equivalent to maximizing $h(\tau)=\lambda(\tau)(T-\tau)$, which is strictly concave because $h''=\lambda''(T-\tau)-2\lambda'<0$. Since $h'(0)=a\Delta\lambda T-\lambda_0$ and $h'(T)<0$, the optimum satisfies
\begin{equation}
\tau^*=0\quad\text{if}\quad T\leq T_{\mathrm{crit}}:=\frac{\lambda_0}{a\,\Delta\lambda},
\qquad
a\Delta\lambda\,e^{-a\tau^*}(T-\tau^*)=\lambda(\tau^*)\quad\text{otherwise},
\label{eq:tau-opt}
\end{equation}
with a unique interior solution in the second case. Under this model, a budget threshold therefore arises, with no adaptation fraction fixed in advance: above $T_{\mathrm{crit}}$ a strictly positive adaptation stage is optimal, and below it direct training is. Implicit differentiation of the interior condition gives
\begin{equation}
\frac{d\tau^*}{dT}=\frac{1}{a(T-\tau^*)+2}\in\Big(0,\tfrac12\Big),
\label{eq:tau-growth}
\end{equation}
so the optimal adaptation length grows by less than half of each additional update. Its fraction $\tau^*/T$ first rises from zero at $T_{\mathrm{crit}}$ and decreases only where $d\tau^*/dT<\tau^*/T$; sublinear growth alone does not imply a decreasing fraction. The experiments evaluate a 3,000-update stage but do not fit or validate this optimal-length curve; at a fixed budget, the optimal length is set by the rate at which adaptation saturates, $a$, relative to the base rate $\lambda_0$.

\subsection{Token Accounting}
\label{sec:tokens}
Let $U$ be the total updates, $s$ the S1 updates, $b$ the batch size, and $t_w,t_a$ the video-model and action tokens per example. With the same video-model sequence length in both schedules, the transformer input-token counts are
\begin{align}
\mathcal T_D&=bU(t_w+t_a),\nonumber\\
\mathcal T_S&=b\{s\,t_w+(U-s)(t_w+t_a)\},\nonumber\\
\mathcal T_D-\mathcal T_S&=b\,s\,t_a.
\label{eq:tokens}
\end{align}
Both schedules process the same number of video-model tokens. Staging omits action tokens during the $s$ adaptation updates. For $s/U=3,000/30,000$, this reduces action-token exposures by 10\%, equivalent to $0.1\,t_a/(t_w+t_a)$ of the combined world and action token count.

\begin{figure}[!htbp]
\centering
\paperfigure{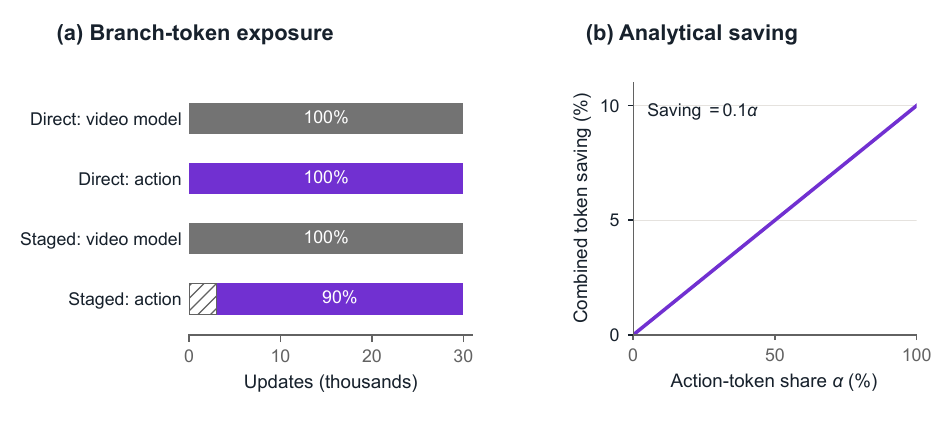}
\caption{\textbf{Video- and action-token accounting.} (a) Both schedules process video-model tokens for 30k updates; staging processes action tokens for 27k. Gray denotes video-model and purple action tokens. Labels give exposure relative to direct training; hatching marks omitted action work. (b) The analytical identity $0.1\alpha$ for $\alpha=t_a/(t_w+t_a)$, where $t_w,t_a$ count tokens per example. Values come from token counts.}
\label{fig:token-accounting}
\end{figure}

\cref{fig:token-accounting} expresses the reduction as a function of the action-token share. \cref{eq:tokens} compares token counts at a fixed update budget, whereas \cref{eq:target-budget} compares budgets at a fixed performance target.

At a fixed token budget $B_{\mathrm{tok}}\geq b s t_w$, direct training permits $U_D=B_{\mathrm{tok}}/[b(t_w+t_a)]$ updates. Allocating $b\,s\,t_w$ tokens to S1 leaves a total staged-update budget of
\begin{equation}
U_S=s+\Big(U_D-s\tfrac{t_w}{t_w+t_a}\Big)=U_D+s\tfrac{t_a}{t_w+t_a}>U_D
\label{eq:equal-token}
\end{equation}
for $s>0$ and $t_a>0$. The additional updates result from replacing some joint updates with video-model-only updates.

\subsection{Training-Cost Decomposition}
\label{sec:theory-cost}
Training cost can be separated into update count and per-update cost. Let the direct and staged schedules use $U_D$ and $U_S$ updates, with $s$ S1 updates in the staged schedule. Assume constant per-update costs $c_1$ and $c_2$ for S1 and S2. Then $C_D=c_2U_D$, $C_S=c_1s+c_2(U_S-s)$, and
\begin{equation}
C_D-C_S=\underbrace{c_2\,(U_D-U_S)}_{\text{fewer updates}}+\underbrace{(c_2-c_1)\,s}_{\text{S1 cost difference}}.
\label{eq:cost-split}
\end{equation}
The first term accounts for the difference in update count; the second accounts for the per-update cost difference during S1. Let $\rho=c_1/c_2$. For the observed staged 13,000-update and direct 30,000-update comparison, the modeled cost ratio is
\begin{equation}
\frac{C_S}{C_D}=\frac{3\mathrm{k}\,\rho+10\mathrm{k}}{30\mathrm{k}}=\tfrac13+0.1\rho.
\label{eq:cost-ratio}
\end{equation}
For $0\leq\rho\leq1$, the ratio ranges from 33.3\% to 43.3\%. Against the 50,000-update direct checkpoint, which the staged 13,000-update checkpoint also exceeds, the ratio is $(3\rho+10)/50$, ranging from 20.0\% to 26.0\%. The update ratios reported in \cref{sec:res-crossover}, 43.3\% and 26.0\%, therefore correspond to the endpoint $\rho=1$, which assigns each adaptation update the full cost of a joint update; if the omitted policy and geometry pathways make $\rho<1$, the modeled cost ratio is lower. The actual per-update cost ratio is not measured. At matched 30,000-update budgets, the corresponding ratio is $0.9+0.1\rho$, ranging from 90\% to 100\%.

The cost of adaptation is also amortized when the adapted video generation model is reused. Suppose one S1 checkpoint serves $M$ downstream tasks, each requiring $k_S$ S2 updates on the adapted video generation model or $k_D$ direct updates without it. Shared adaptation costs $c_1s+Mc_2k_S$, whereas independent direct training costs $Mc_2k_D$, so whenever $k_D>k_S$ reuse has lower total cost as soon as
\begin{equation}
M>\frac{c_1\,s}{c_2\,(k_D-k_S)}=\frac{\rho\,s}{k_D-k_S}.
\label{eq:reuse}
\end{equation}
With $s=3{,}000$ and the measured pair $k_S=10{,}000$ S2 updates against $k_D=30{,}000$, the threshold is $M>0.15\rho$: under the assumed $0\leq\rho\leq1$, one task with these update counts recovers the adaptation cost. If the same adapted video generation model provides the assumed per-task savings on additional tasks, its adaptation cost can be amortized according to \cref{eq:reuse}, adding $c_2(k_D-k_S)$ in savings per task. This cross-task benefit is a model prediction.

\cref{eq:cost-split} separates the two efficiency benefits of \cref{sec:res-crossover}. The first term uses the update-count difference between the compared checkpoints; the second uses an assumed per-update cost difference. The token identity in \cref{eq:tokens} counts omitted inputs and does not imply proportional FLOP or runtime savings.

\ReportAppendixSection{Pretraining Data Details}{app:data}

\subsection{Cleaning and Selection Overview}
The pretraining corpus consists of real-robot recordings selected after cleaning: 6,000 hours of ZERITH H1 data and 4,000 hours of public data, totaling 10,000 hours (\cref{tab:data-composition}). The following sections describe cleaning and repair; their intermediate processing statistics are distinct from the final selected training volumes.

The dataset inventory, preprocessing output, and final training selection use different units and scopes. An ingested MCAP file is counted as one data record, while cleaned hours measure the retained recording duration. Multiple camera streams are components of a record. Counts and hours are reported separately because filtering, temporal trimming, and repair can change duration without a proportional change in record count.

\subsection{Quality-Control and Repair Procedure}
Processing consists of record parsing, quality checks, repair, conversion to a common training format, and joint filtering. Checks examine stationary segments, within-trajectory stationary intervals, non-image stream gaps, camera frame loss, and timestamp consistency. Most checks assign \texttt{pass}, \texttt{repairable}, or \texttt{error}; camera checks use a binary result. A non-image gap of at most 14 frames is classified as repairable, whereas a longer gap is rejected under the reported rule.

Repair operations include end-effector coordinate alignment, temporal trimming, interpolation of eligible gaps, and forward-kinematics recomputation. ZERITH H1 kinematics are expressed in a common body reference frame. Gripper commands are converted from source-specific normalized or metric representations to a consistent convention. Additional diagnostics inspect end-effector jerk and joint singularities. Final filtering combines quality labels by record identifier. The per-check percentages describe overlapping sets of records.

\subsection{ABC-130K Processing}
The ABC-130K paper reports 134,806 episodes, 195 tasks, and 3,553 hours~\cite{abc130k}. \cref{tab:abc-qc} gives the reported per-check quality distribution. Non-image stream gaps have the largest repairable fraction, 73.01\%, and an error rate of 5.08\%. Timestamp overlap has a 30.63\% repairable fraction and a 2.81\% error rate. The camera check has a 97.40\% pass rate; this check measures frame completeness.

\begin{table}[htb]
\centering
\caption{\textbf{ABC-130K quality-check outcomes over 130,688 ingested records.} Percentages are calculated separately for each check. A dash indicates an unreported or inapplicable category.}
\label{tab:abc-qc}
\small\setlength{\tabcolsep}{8pt}
\begin{adjustbox}{max width=\linewidth}
\begin{tabular*}{0.80\linewidth}{@{\extracolsep{\fill}}lrrr@{}}
\toprule
\textbf{Check} & \textbf{Pass (\%)} & \textbf{Repairable (\%)} & \textbf{Error/fail (\%)}\\
\midrule
Non-image stream gaps & 21.91 & 73.01 & 5.08\\
Timestamp overlap & 66.55 & 30.63 & 2.81\\
Stationary segments & $\approx99.94$ & 0.06 & ---\\
Camera frame loss & 97.40 & --- & 2.60\\
\bottomrule
\end{tabular*}
\end{adjustbox}
\end{table}

Filtering alone retained 2,363 hours. Interpolation and alignment repair increased the retained duration to 2,864 hours, an increase of 501 hours (21.2\%) recovered from existing recordings. The repaired output is one source for the final public-data selection in \cref{tab:data-composition}.

\subsection{Combined Public Sources}
RoboMIND 2.0, AgiBot World Beta, and RoboChallenge are processed together. Quality checks identified numerical outliers and frame gaps requiring repair. Approximately 1,400 hours were retained after processing.

Processing applies end-effector alignment and temporal trimming, and checks stream gaps with automatic detection and a threshold multiplier of 1.5. The filtering rule requires the stationary-segment check to be pass or repairable, and both the stream-gap and camera checks to pass. The rule accepts all three timestamp labels after the alignment and trimming stage. The acceptance rule therefore distinguishes the timestamp detector output from subsequent repair operations.

Dataset-specific issues included camera--action misalignment, long stationary segments, and joint singularities in RoboChallenge; corrupted HDF5 files, zero poses, duplicated state and action values, and control delay in RoboMIND; and action-stream gaps, inconsistent gripper units, and approximately one-second delay in AgiBot Beta. These findings motivate checks for coordinate consistency, temporal alignment, numerical validity, and state--action duplication before training. The combined processed subset contains approximately 1,400 hours.

The public pretraining selection combines ABC-130K with the jointly processed sources above and contains 4,000 hours after cleaning. Intermediate processing durations and the final selected training volume are reported as different stages.

\subsection{ZERITH H1 Recordings}
After cleaning, we select 6,000 hours of our ZERITH H1 real-robot recordings for pretraining. The data follow the quality-control and repair procedure described above.

\subsection{Source Inventory}
\cref{tab:data-inventory} lists ingested records by source.

\begin{table}[htb]
\centering
\caption{\textbf{Ingested MCAP records by source.} Counts are taken before filtering.}
\label{tab:data-inventory}
\small\setlength{\tabcolsep}{8pt}
\begin{adjustbox}{max width=\linewidth}
\begin{tabular*}{0.80\linewidth}{@{\extracolsep{\fill}}lrP{.39\linewidth}@{}}
\toprule
\textbf{Source label} & \textbf{Records} & \textbf{Inventory category}\\
\midrule
ABC-130K & 130,688 & Public source\\
AgiBot World Beta & 158,927 & Public source\\
RoboMIND 2.0 & 123,489 & Public source\\
RoboChallenge & 32,939 & Public source\\
\bottomrule
\end{tabular*}
\end{adjustbox}
\end{table}

Published source sizes provide context for the ingestion records. RoboMIND 2.0 reports more than 310,000 trajectories across six embodiments and 739 tasks~\cite{robomind2}; AgiBot World reports more than one million trajectories across 217 tasks~\cite{agibot}. These source totals differ from the subsets processed here. The ABC-130K statistics above refer to its paper release; the dataset card lists 130,703 YAM episodes and 3,590.7 hours~\cite{abccard}.

\ReportAppendixSection{Implementation Details}{sec:practice}
The video-model transformer uses width 3072 and 24 attention heads of dimension 128; policy blocks match this configuration. LIBERO inputs use front and wrist camera views, while the real robot uses head and left/right wrist views. Robot joint and gripper states carry per-dimension validity masks.

\subsection{Gradient Computation and Synchronization}
Gradient norms, clipping, and finite-value checks are computed over the optimizer's parameter groups rather than over the model's parameters. Under sharded mixed-precision training this choice is what makes the norm exact: the framework synchronizes each gradient and moves it into an FP32 shard owned by the optimizer, so the optimizer groups are the place where the complete applied gradient resides. Clipping and telemetry operate on optimizer-managed gradients, not on the AdamW parameter update itself. The released trainer records its global norm before clipping; this does not establish the aggregation or clipping convention of the separate probes in \cref{sec:probes}. Finite-value checks are fused into the global norm reduction. A non-finite local loss or norm is encoded as NaN before the single sum-reduction that computes the norm; because NaN propagates through the sum, every worker observes the same non-finite total and raises a synchronized error before the optimizer step; the failed update is not applied.

The routing schedule is defined relative to the S2 update count, so its absolute length scales with the stage. A 30,000-update S2 run uses 1,500 updates at $\beta=0$ followed by a 1,500-update ramp, and a 27,000-update S2 run uses 1,350 updates for each interval. Visual-loss gradients remain active throughout; the schedule governs only the action gradient's access to the video generation model, and the video generation model continues to learn from its visual objective during the interval at $\beta=0$.

\subsection{Precision, Masks, and Feature Caching}
The backbone runs in BF16, while action normalization, statistics, and time-sensitive operations remain in FP32, so the quantities that define the action space are never accumulated at reduced precision. Invalid action dimensions, missing camera slots, and invalid state entries are represented as zeros paired with false validity masks. False validity masks exclude entries from the losses and attention operations to which they are applied; zero filling alone does not make an entry inert in every computation. Finite placeholders avoid propagating NaN or infinity into losses or normalizers. The losses exclude masked entries, and every model input is verified to be finite. Geometry features are precomputed under a fixed sampling seed and read from a cache, so the frozen geometry encoder is not re-executed during training. This fixes the cached features but does not guarantee identical numerical behavior across runs. The configuration fits in accelerator memory directly, so it runs with gradient checkpointing, CPU offload, and graph compilation disabled.

\subsection{Monitoring and Reproducibility}
Training health is assessed from four signals read together: per-module gradient norms, the visual and action losses reported separately, the full pure-noise sampling error, and closed-loop success. Each signal places the others in context. A gradient norm is recorded together with whether it precedes or follows clipping, since the two differ by construction once clipping is active, and action errors are expressed in normalized action coordinates. Closed-loop success is the primary criterion, because the ten-step solved trajectory that the controller executes is a different object from a one-step reconstruction of the same target. The noise comparison of \cref{sec:res-noise} illustrates this: both S1 variants recover recognizable scene structure in the frame-16 examples, yet their reported rollout success differs by 20.80 points.

Each run binds its checkpoint, normalization statistics, dataset path, source revision, device configuration, and batch size into a frozen record, and both checkpoint restoration and export verify that record directly. Export compares the model's codec statistics with the target statistics buffer by buffer and accepts a checkpoint only when every buffer matches. These checks verify the recorded fields on restoration and export; they do not establish equality of unrecorded run settings or eliminate all numerical variation.

\ReportAppendixSection{Decoded Visual Probes}{app:qualitative}

\begin{figure}[!htbp]
\centering
\paperfigure{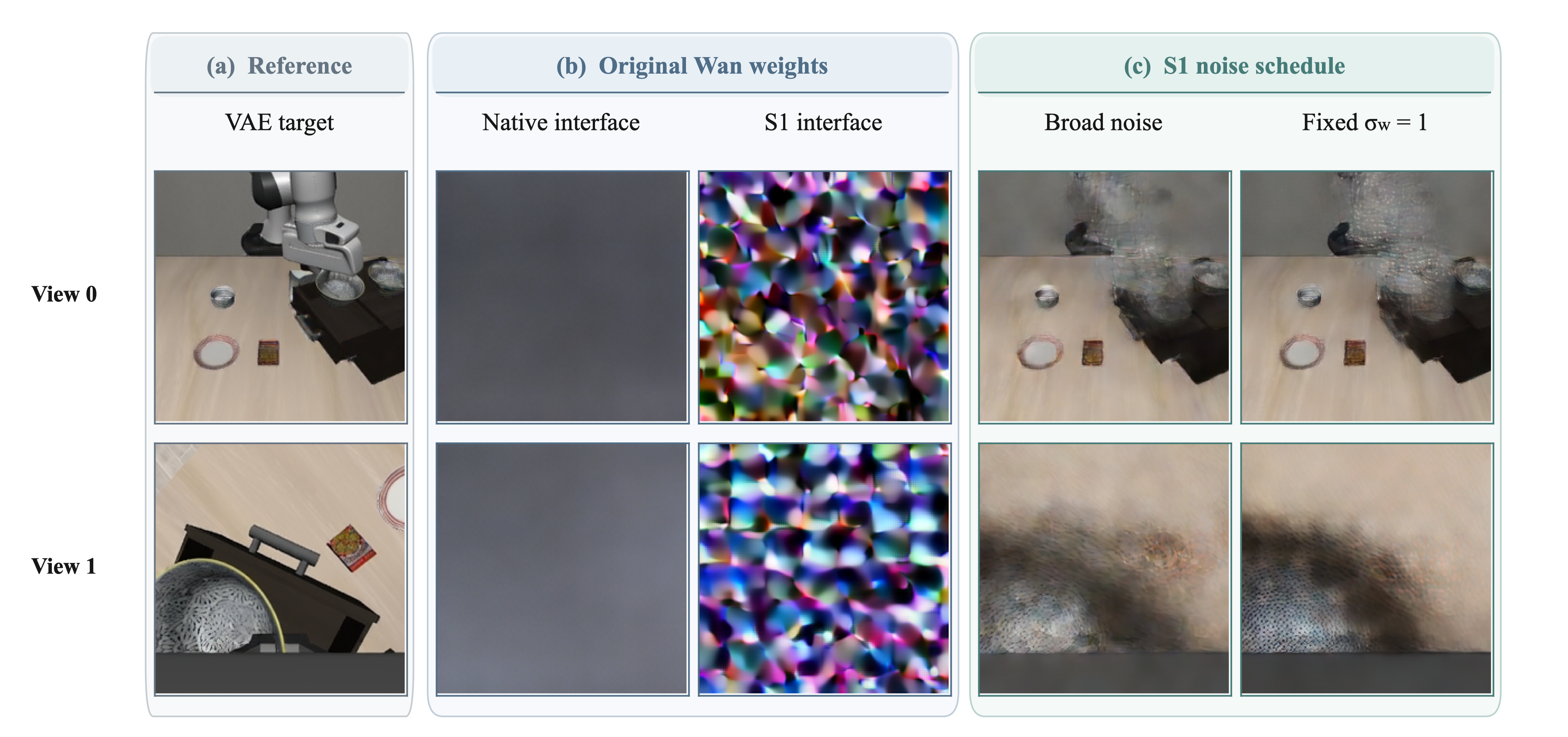}
\caption{\textbf{Qualitative visual predictions.} Offline frame-16 decoding, with two camera views. Columns: VAE target, original Wan with its native interface, original weights with the S1 interface, broad-noise S1, and fixed-$\sigma_w=1$ S1. Both S1 variants recover recognizable scene structure; the downstream noise control in \cref{tab:noise-controls} distinguishes their control performance.}
\label{fig:qualitative}
\end{figure}

\cref{fig:qualitative} decodes frame 16 offline for two camera views under five conditions and separates two effects that the loss probes report jointly. The third column applies the S1 interface to the original weights, isolating the change of interface from the change of weights; the fourth and fifth columns use the same interface and show broad-noise and fixed-noise S1. Both adapted variants recover recognizable scene structure, although their downstream LIBERO-Plus success differs by 20.80 points. These single-horizon examples do not quantify reconstruction quality or establish equivalence between the schedules; closed-loop success is evaluated separately, as in the noise comparison of \cref{sec:res-noise}.

\ReportAppendixSection{Extended Limitations}{app:limitations}

\subsection{Statistical Precision}
The staging, dataset, and geometry-operator studies report three-seed means, and their principal paired comparisons report the sample SD of seed-wise differences. SD bars describe variability across seeds. The branch-initialization, noise-control, learning-curve, budget-sweep, attention, and pretraining results are single runs: they carry scenario-level sampling error, but we have no estimate of their seed-to-seed variation. On the 10,030-scenario LIBERO-Plus manifest, the binomial standard error of a difference between two independent scores near 40\% is about 0.7 points. This figure treats the two runs as independent samples; a paired test would need per-scenario outcomes, which we do not report. At this precision, the 0.51-point LIBERO-Plus advantage of $R_0$ over $R_1$ at $W_1$ ($z\approx0.7$) is within noise, and standard LIBERO favors $R_1$ by 0.65 points ($z\approx1.0$). We deploy $(W_1,R_0)$ because it has the higher recorded LIBERO-Plus score.

After S1, the multiplicative-versus-static difference is $0.68\pm0.30$ points (mean $\pm$ paired SD), and the staging--conditioning interaction has no SD of its own. The 4.6--6.1-fold ratio between effects on LIBERO-Plus and standard LIBERO is a ratio of point estimates. Policy-interface comparisons combine a single deployed score with three-seed alternatives. Comparisons are made within each experimental group and pretraining background, and groups with similar baselines are kept separate. The mixed-data extensions change both the objective and the pretraining exposure, so their 4.12-point difference mixes the effect of action supervision with that of extra compute.

\subsection{Crossover and Training Savings}
At a matched 5,000 total updates, direct training leads by 17.06 points, about 37 standard errors. Past that budget the evidence is weaker, because direct training was not evaluated at the intermediate budgets. The staged 8,000-update checkpoint exceeds the direct 10,000-update checkpoint by only 0.14 points (0.2 standard errors). The staged 13,000-update checkpoint exceeds the direct 15,000-update checkpoint by 6.19 points ($z\approx9$), but this still compares unequal budgets. Both the 5,000--8,000-update interval and the broader 5,000--13,000-update interval ($1.7\tau$--$4.3\tau$) therefore rest on assumptions about unmeasured direct scores. Early matched-S2 comparisons are mixed: at 1,000 S2 updates the staged schedule trails by 1.44 points, and the three earliest differences all lie within 1.5 points at success rates below 5\%.

The savings in \cref{tab:matched} pair reference checkpoints on a coarse grid of single runs. On such a grid, each checkpoint only bounds from above the updates its schedule needs to reach a given level, so the savings are not minimum-hitting-time estimates. The table's 500-update row compares matched S2 updates rather than matched totals. The staged 13,000-update checkpoint exceeds the best direct score, 38.52\%, by 1.34 points ($z\approx1.9$). The first staged checkpoint more than four standard errors above that score is at 18,000 updates, 3.30 points higher, which uses 36\% of the 50,000-update direct budget and saves 10.24M sample exposures. The direct curve is non-monotone over the eleven evaluated budgets and could reach a given level at an unevaluated budget. Between 30,000 and 50,000 updates the staged advantage narrows by 1.24 points per 10,000 updates; this two-endpoint slope is too coarse to extrapolate. The staged score also declines after its 35,000-update peak, so both the advantage beyond 50,000 updates and the location of the staged optimum remain open.

The crossover analysis uses one adaptation length, one benchmark suite, and one budget axis, update count; token-exposure and wall-clock axes are left for future work. Exposure savings count repeated passes over the same 1,693 demonstrations. Because neither the number of demonstrations nor the batch size is varied, we cannot estimate a sample threshold $N^*$ or say whether fewer demonstrations would suffice.

\subsection{Theoretical Models}
\cref{app:theory} states the assumptions of each analysis alongside it. This subsection lists the quantities those analyses rely on that we do not measure, and what follows from them.

\emph{Noise coverage.} The noise-level analysis concerns population optima. It shows that the optimal output at $\sigma_w=1$ depends only on the conditional mean; whether endpoint-trained features carry only that information, and whether broad-noise training improves the policy's representation, are separate questions. Our S1 distribution has no atom at $\sigma_w=1$, so the endpoint statement applies to it only by continuity. $J_\sigma$ and $G_p$ are unmeasured. We report no per-category LIBERO-Plus breakdown, so the prediction that broad noise helps most on samples with uncertain futures remains untested.

\emph{Tracking and transfer.} The tracking bound assumes local strong convexity, smoothness, bounded mixed curvature, and deterministic gradient descent, whereas the policy is trained with AdamW. The constants $\mu$, $L$, $M$, and $K$ and the video-model parameter drift during S2 are unmeasured. Raw gradient norms do not set AdamW step sizes, so the 95.2\% reduction in the initial visual-loss gradient need not translate into a 95.2\% reduction in video-model updating. The branch-transfer analysis is exact only where the policy objective is quadratic around both $R_0$ and $R_1$; $g_i$, $H_i$, and $H_{rw}$ are unmeasured, and the interpolation optimum $\lambda^*$ is derived but not evaluated. The spectral account of the policy-only result fits two observations, the initial action loss and the final success rate, and other mechanisms could fit them as well.

\emph{Budget and cost models.} The linear crossover model assumes constant stage-wise rates; the observed plateau and narrowing depart from it over the full budget range. Setting $r_{AC}=0$ assumes a control metric that is defined during policy-free S1, and we evaluate no such probe. The rate ratio of 1.6--2.5 and limiting saving of 37.5--60\% follow from the 5,000--8,000-update interval; the 5,000--13,000-update interval gives 1.3--2.5 and 23--60\%. The optimal-length model is illustrative: no parameter is fitted and only $\tau=3{,}000$ is evaluated, so whether 3,000 updates is close to optimal remains open. The per-update cost ratio $\rho$ is untimed, so the 33.3--43.3\% and 20.0--26.0\% cost ranges are model consequences of measured update counts rather than measured GPU time. The token identity counts transformer input tokens only, and we did not run the equal-token comparison of \cref{eq:equal-token}. The reuse threshold assumes that one adapted video generation model serves every downstream task equally well; with a single adaptation domain evaluated, cross-task amortization remains a property of the cost model.

\subsection{Mechanistic Interpretation}
Three interpretations in the main text fit the measurements but have not been tested by intervention: that copied S1 layers carry a visual-prediction specialization that does not suit the policy, that S2 begins in a state in which action updates are more effective, and that the larger relative response to cross-camera masking reflects the value of cross-view information. ``Specialization'' here means functional specialization only; testing it would require representation-level probes across checkpoints. The initialization probes use eight batches at the start of S2 and describe only that point. The near-zero gradient cosines argue against gradient conflict as the mechanism of the staging effect.

The fixed-noise S1 and full-isolation direct controls are separate single-run comparisons. Broad-noise S1 followed by fully isolated S2, and intermediate noise distributions, are not evaluated, so we cannot say whether each ingredient is independently necessary. The larger video-model adaptation effect on LIBERO-Plus compares aggregate suite scores, without a breakdown by perturbation category. On standard LIBERO, mixed-data S1 ranks above target S1 and multiplicative conditioning is slightly negative, so those conclusions apply to the perturbation suite.

Attention statistics are aggregated over the tapped layers and three denoising steps without per-layer or per-anchor dispersion, so their differences have no intervals. Attention mass measures where the policy looks, which need not be what the action prediction depends on. The 51.8\% drop in geometry attention and the staging--conditioning interaction come from different interventions. Establishing causal roles would require masking or permuting action tokens, past frames, or dynamic-video tokens under matched inputs. The decoded frames are offline single-horizon examples.

The policy-only result also has a practical implication: a checkpoint that is better on the visual objective can still be a poor initialization for another module. Its explanation in \cref{sec:theory-transfer} remains a hypothesis.

\subsection{Data and Implementation}
Data statistics come from the processing pipeline and were not re-derived from raw recordings. The selected data volumes do not identify which data each checkpoint consumed, and the effect of individual repair operations on policy success is unmeasured.

Several choices in \cref{sec:system,sec:method,sec:practice} rest on design reasoning rather than ablation. Layer selection was compared across three configurations, with differences of 0.15--0.20 points, and we report no per-layer representation probe. The length of the $\beta=0$ interval and the ramp of the routing schedule are fixed by design, and we do not measure when the policy begins to produce useful gradients. Retaining $\mathcal L_w$ in S2 is never compared against dropping it. Whether the binding gates move away from zero during training is unreported. The precision split, masking convention, disabled memory optimizations, and norm computation over optimizer groups were each chosen without comparing alternatives. Run records verify the recorded configuration but cannot capture every source of variation, such as nondeterministic kernels. All runs use four nodes with eight accelerators each and no gradient accumulation; because data quotas are tied to the global batch size, a single-node schedule with accumulation may behave differently.

\subsection{Evaluation and Deployment}
Baseline evaluation episodes, seeds, and fine-tuning budgets differ across systems, so the margins in \cref{tab:comparison} compare reported numbers. The MT50 references aggregate four difficulty groups with equal weight, whereas our score pools episodes, so that margin is a cross-study comparison. We report no per-suite or per-task outcome counts or evaluation-repeat details for the pretraining sweep and the MT50 result. Complete-system scores include robot-data pretraining and therefore isolate neither the staged schedule nor the policy initialization. Each pretraining checkpoint is a single run, and the standard-LIBERO column is non-monotone.

Hardware evaluation reports task-level success on pick and place and bin lining, comparing complete systems. Sorting is shown only as a qualitative example; wrist paths are manual image-plane annotations, and timestamps refer to video playback. Inference latency is measured for the deployed configuration on one H800 at batch size one, so it describes neither throughput nor the baselines.

\ReportLabFooter

\end{document}